\documentclass{article}
\usepackage{iclr2027_conference,times}
\usepackage{amsmath,amssymb,graphicx}
\usepackage{fontawesome5}
\usepackage{hyperref}
\usepackage{url}
\usepackage{capt-of}
\usepackage{longtable}
\usepackage{placeins}
\usepackage{array,multirow}
\hypersetup{hidelinks}

\vspace{-20pt}
\title{Confidence-Ordering Reversal under\\Contextual Priors in Neural Decoding}

\iclrfinalcopy
\author{
Xinyu Zhang \textsuperscript{1,2$\dagger$}
\hspace{1.1em}
Sichao Liu\textsuperscript{1$\dagger \ast$}
\\[1.5ex]
\textsuperscript{1}~KTH \quad
\textsuperscript{2}~Karolinska Institutet
}
\begin{document}
\maketitle
\begingroup
\renewcommand\thefootnote{}
\footnotetext{\textsuperscript{$\dagger$} equal contributors; \textsuperscript{$\ast$} Corresponding author: Sichao Liu (\url{sicliu@kth.se})}
\endgroup

\begin{abstract}
Contextual priors improve neural-to-language decoding by reshaping candidate scores. However, confidence is read from the same reshaped scores, so the errors a prior leaves behind can become more confident with no change in accuracy to reveal it. In this work, we ask how a prior shapes the confidence of these errors, studying speech retrieval on MEG-MASC and MOUS with local decoding scores and a contextual prior combined by additive shallow fusion, and the fused top-two margin as confidence. Among initially incorrect predictions, we find a \textbf{confidence-ordering reversal}: a larger margin makes a \emph{repair}, an error that fusion corrects, more likely when the correct candidate starts near the top of the local ranking, but less likely when it starts lower. On MEG-MASC, pooled correctness AUROC is 0.87, yet the AUROC separating repairs from \emph{residual errors}, which fusion leaves uncorrected, \textbf{falls from 0.70 at initial ranks 2–3 to 0.39 at ranks 21–50}. Errors starting beyond rank 20, inside the reversed region, make up 46.6\% of all errors after fusion. We propose a score-level account: a repair must first close the correct candidate's initial deficit, which limits its final margin, whereas a residual error can build a large margin between two incorrect candidates. Through a causal intervention that changes only the fusion weight, \textbf{we show that the reversal moves to deeper ranks, as the account predicts}, and that under a word-level LM prior it keeps moving after accuracy gain peaks, so a weight chosen for accuracy does not settle confidence. Guided by this account, we read local and prior scores separately: read before fusion, the prior's own scores already separate repairs from residual errors where the fused margin reverses, and estimators built on local and prior scores let a selective decoder answer on 74.5\% of windows instead of 56.7\%, with 92\% of its output sets still containing the correct candidate. \textbf{Confidence after contextual fusion should retain the local and contextual evidence behind each prediction, not just the fused scores}. \noindent
\href{https://confidencereversal.github.io/}
{\textcolor{blue}{\faGlobe\ Project Page}}
\quad
\href{https://github.com/AmadeusFake/NeuDecodingConfReversal}
{\textcolor{blue}{\faGithub\ Code}}
\end{abstract} 

\section{Introduction}
\label{sec:introduction}

Contextual priors can improve neural-to-language decoding by supplying linguistic structure and sequential context that local neural predictions can miss~\citep{willett2023high,d2025towards}. They also change the remaining errors. Recent MEG work reports fluent sentences whose content differs from the intended text when neural evidence is weak~\citep{zhang2026brain2qwertyv2}, and invasive brain-to-text systems exhibit confident but incorrect neural predictions~\citep{huang2026probabilistic}. In sustained BCI use, users must identify and correct decoded errors~\citep{speier2016integrating,card2026longterm}; when confidence determines which outputs are flagged for review, a confidently scored error can pass unflagged and be left for the user to catch. Evaluation nonetheless centres on decoding accuracy or text similarity~\citep{silva2024speech}. When confidence guides selective output~\citep{kim2026confidence}, the prior shapes both the prediction and the fused scores used to judge it.

\begin{figure}[ht]
\centering
\includegraphics[width=\linewidth]{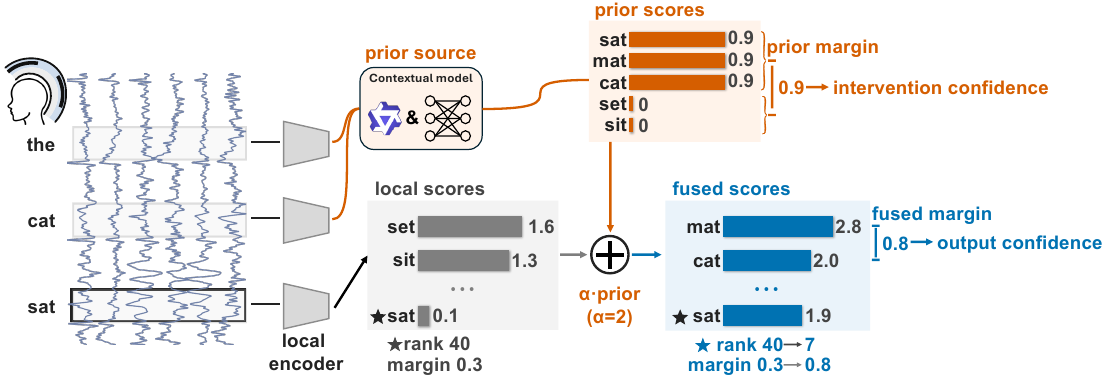}
\caption{\textbf{Contextual fusion can make an error more confident.} A local encoder scores candidates from the current MEG window, while a contextual prior from preceding windows adjusts these scores through additive fusion with weight $\alpha=2$. The prior operates at a coarser granularity, so several candidates share one score. Fusion advances the correct candidate ($\star$) from rank 40 to rank 7 without correcting the prediction, while increasing the top-two margin from 0.3 to 0.8. Brackets mark the prior and fused margins. Values are illustrative; real cases are in Table~\ref{tab:examples}.}
\vspace{-0.5em}
\label{fig:overview}
\end{figure}

We find that strong overall confidence discrimination can conceal opposite orderings of correction outcomes. We study this in offline speech retrieval~\citep{defossez2023decoding}, retaining a fixed candidate pool and separate local and prior scores (Figure~\ref{fig:overview}). The \emph{fused margin}, the gap between the two highest fused scores, serves as a confidence measure. Among initially incorrect predictions, \emph{repairs} become correct after fusion and \emph{residual errors} remain incorrect. Across MEG-MASC~\citep{gwilliams2023meg} and MOUS~\citep{schoffelen2019204}, larger margins favour repairs when the correct candidate starts near the top of the local ranking, but residual errors when it starts lower (Figure~\ref{fig:reversal}). On MEG-MASC, pooled correctness AUROC is 0.87, yet repair-separation AUROC falls from 0.70 at ranks 2--3 to 0.39 at ranks 21--50. We call this a \textbf{confidence-ordering reversal}. The reversed region is where errors concentrate: residual errors starting beyond rank 20 account for 46.6\% of all post-fusion errors on MEG-MASC. The pre-fusion margin favours residual errors in every group, so fusion itself produces the positive ordering near the top. \textbf{How does a prior make some errors more confident, and what information can still help identify them?}

\textbf{We trace the reversal to how repairs and residual errors form their margins.} A repair must first close the correct candidate's initial score deficit, so its lead is bounded by the remaining contextual advantage, whereas a residual error can acquire a large lead between two incorrect candidates even as the correct candidate advances. Because this bound follows from additive fusion, it holds wherever a prior is added to local scores, and it predicts that increasing fusion weight extends positive repair ordering toward larger initial deficits.

We test this prediction with a causal intervention that varies only the fusion weight, keeping local and prior scores fixed. Under the history-aggregation prior, built from preceding windows, positive ordering extends from no rank group to ranks 11--20 and to larger score deficits, and the boundary never retreats in participant-bootstrap resamples (Figure~\ref{fig:reach}). Under a word-level teacher-forced language-model prior, it also expands down the initial ranking and continues after net accuracy gain has peaked: \textbf{accuracy and confidence ordering are two different responses to the same prior.}

Grouping errors by initial rank exposes the reversal, and the margin-formation account locates its source: fusion merges local competition and prior support into one margin. Guided by this account, we read the two scores separately: the prior's own margin separates the outcomes that the fused margin reverses (Figure~\ref{fig:reversal}C), and learned estimators built on both recover positive repair ordering. Feature comparisons show that local features improve selective output beyond richer fused-score features, with further gains from prior features. At matched 92\% coverage, the full estimator emits on 74.5\% of test windows versus 56.7\% for the fused margin. \textbf{Local and prior scores retain information for identifying confident errors that the final margin does not express.}

This paper proceeds in three steps.

\textbf{Revealing where confident errors arise.} In two MEG datasets, the fused margin orders repairs and residual errors oppositely by initial rank, even though pooled correctness AUROC stays high.

\textbf{Explaining and testing causally.} A score-level account of margin formation explains one way priors make errors more confident; a causal intervention on the fusion weight alone confirms the predicted extension of positive ordering, which accuracy gain does not track.

\textbf{Identifying confident errors from retained information.} Learned estimates recover positive repair ordering, and local and prior features improve selective output beyond richer fused-score features.

\section{Related Work}
\label{sec:related}

\textbf{Contextual priors and the errors they leave behind.}
Contextual priors support BCI communication~\citep{speier2016integrating,herff2015brain}, semantic reconstruction~\citep{tang2023semantic}, and non-invasive decoding with sentence-level context or language-model assistance~\citep{d2025towards,brain2qwerty2026}. When neural evidence is weak, contextual support can dominate the final choice. Reviews and attribution studies already separate neural contributions from contextual gains~\citep{silva2024speech,zhang2026audit,gorenshtein2026influence}. Correctness alone does not show how much neural evidence contributed, and the remaining errors take characteristic forms: non-invasive systems can produce fluent but incorrect text~\citep{zhang2026brain2qwertyv2}, while invasive decoders exhibit overconfident neural predictions~\citep{huang2026probabilistic}. Evaluating contextual assistance therefore also requires assessing whether its remaining errors are identifiable.

\textbf{Confidence formation under contextual priors.}
Confidence already guides selective EEG output~\citep{kim2026confidence}, language-model assistance~\citep{huang2026probabilistic}, and error detection after P300 fusion~\citep{gorenshtein2026trialadaptive}. In ASR, however, shallow fusion can improve recognition while degrading confidence~\citep{kannan2018analysis,li2021confidence,futami2021asr}. Beam analysis connects confidence to local and sequence-level competition~\citep{jia2026beam}, while selective classification reveals group differences concealed by aggregate performance~\citep{jones2020selective}. These observations share a source: under additive fusion, the prior changes both candidate competition and the scores that judge the result. What remains unclear is how confidence orders correction outcomes across groups of initial errors, and how that ordering changes with prior contribution.

\textbf{Judging outputs and contextual revisions.}
In deployment, BCI users correct decoded outputs~\citep{card2026longterm}, and shared-control systems arbitrate between neural commands and machine assistance~\citep{copilot2025,sharedcontrol2019}. For such decisions, selective prediction and prediction sets formalise accepting, expanding, or withholding outputs~\citep{elyaniv2010foundations,geifman2017selective,romano2020classification}, while adaptive fusion adjusts language-model influence~\citep{variani2022adaptive,unni2022adaptive,gong2022entropy}. For contextual revisions, retaining the local prediction provides the alternative, so confidence must support two judgements: final correctness and improvement over that prediction. Applying these decision rules requires knowing which information in local, prior, and fused scores supports each judgement after the prior has acted.

\section{Method}
\label{sec:method}

\subsection{Local scores and contextual fusion}

Following the speech-retrieval framework of~\citet{defossez2023decoding}, each query is a word-aligned, 3-s MEG window paired with its correct audio segment in a fixed candidate pool $\mathcal Y=\{1,\ldots,O\}$. At window index $t$, the local decoder maps the neural window $x_t$ to scores $s_t\in\mathbb R^O$, while the contextual prior maps preceding history $h_{t-1}$ to scores $\pi_t$ over the same candidates. Let $Y_t$ be the correct-candidate random variable and $y_t$ its evaluation label; $j$ and $k$ index candidates. The distributions $q_t=\operatorname{softmax}(s_t)$ and $p_t=\operatorname{softmax}(\pi_t)$ define the local and prior model probabilities. Additive shallow fusion with weight $\alpha\geq0$~\citep{kannan2018analysis} gives
\begin{align}
\tilde s_{t,j}&=s_{t,j}+\alpha\pi_{t,j},
\label{eq:fusion}\\
\Pr_{\alpha}(Y_t=j\mid x_t,h_{t-1})
&=\frac{q_t(j)p_t(j)^\alpha}{\sum_k q_t(k)p_t(k)^\alpha}.
\label{eq:poe}
\end{align}
Equation~\ref{eq:poe} is the softmax of Eq.~\ref{eq:fusion}, so fusion is a tempered product of the two distributions and reduces to the local one at $\alpha=0$; for $\alpha>0$ it gives the prior temperature $T_{\mathrm{prior}}=1/\alpha$ while leaving local scores unchanged, so $\alpha$ controls the prior's contribution relative to neural evidence (Appendix~\ref{app:algebra}). The local and fused predictions are $\hat y_t=\arg\max_j s_{t,j}$ and $\tilde y_t=\arg\max_j\tilde s_{t,j}$.

For a score vector $a$ with largest entries $a_{(1)}\geq a_{(2)}$, its top-two margin is $m(a)=a_{(1)}-a_{(2)}$, equal to the log odds between its two leading softmax candidates. The \emph{pre-fusion margin} is $m(s_t)$ and the \emph{fused margin} is $m(\tilde s_t)$. The \emph{prior margin} is computed over the prior's own entries, distinct sentences or words, before their scores are shared by the candidates that map to them; entries with equal scores remain separate. All three margins are evaluated by how they order outcomes.

\subsection{Correction outcomes, ordering, and reach}
\label{sec:correction-outcomes}

Among initially incorrect predictions ($\hat y_t\ne y_t$), a \emph{repair} has $\tilde y_t=y_t$ and a \emph{residual error} has $\tilde y_t\ne y_t$; among initially correct predictions, a \emph{regression} has $\tilde y_t\ne y_t$. Let $n_{\mathrm{repair}}$ and $n_{\mathrm{regression}}$ count repairs and regressions over $N$ windows, so that retrieval accuracy changes by $\Delta\mathrm{R@1}=(n_{\mathrm{repair}}-n_{\mathrm{regression}})/N$.

The correct candidate's initial rank is $r_t^0=1+\bigl|\{j:s_{t,j}>s_{t,y_t}\}\bigr|$, the number of candidates scoring strictly above it plus one. We compare repairs with residual errors in fixed groups $\mathcal B$ of initially incorrect predictions, by the correct candidate's initial rank: 2--3, 4--5, 6--10, 11--20, and 21--50, adding 51--100 for the word-level sweeps. Within $\mathcal B$, draw indices $r$ and $e$ uniformly from its repairs and residual errors. Repair-separation AUROC measures whether confidence $z$ ranks the repair above the residual error:
\begin{equation}
\operatorname{AUROC}_{\mathrm{repair},\mathcal B}(z)
=\Pr(z_r>z_e)+\tfrac12\Pr(z_r=z_e).
\label{eq:auroc}
\end{equation}
Values above 0.5 favour repairs; values below 0.5 favour residual errors. We call these positive and negative ordering. Pooled correctness AUROC applies the same expression with $r$ and $e$ drawn from all final correct and all final incorrect predictions. Repair rates across margin quartiles within each group give a complementary view of this ordering. Effective correction reach is the 90th percentile of the initial ranks among repairs:
\begin{equation}
R_{90}=Q_{0.90}\!\left(r_t^0\mid\hat y_t\ne y_t,\ \tilde y_t=y_t\right).
\label{eq:reach}
\end{equation}
Reach summarises the repair distribution independently of confidence scores. Each AUROC curve is described by its deepest positively ordered initial-rank group. For a single downward crossing of 0.5, this locates the reversal boundary; Figure~\ref{fig:reach}C also interpolates the crossing between group midpoints. Full curves describe settings with no observed crossing or multiple sign changes.

\subsection{Datasets and contextual priors}

We evaluate MEG-MASC and MOUS~\citep{gwilliams2023meg,schoffelen2019204} with three independently trained local decoders per dataset (Table~\ref{tab:datasets}). Their neural encoder maps each word-aligned MEG window to a time-resolved speech representation, and candidate scores are inner products with frozen wav2vec2-large-xlsr-53 audio representations~\citep{conneau2021unsupervised}. Stimulus-identity splits evaluate zero-shot retrieval of segments unseen during local-decoder training. Both datasets support the reversal and prior-margin analyses; controlled sweeps, cross-intervention comparisons, and selective-output evaluation use MEG-MASC (Appendix~\ref{app:data}).

\par
\noindent\begin{minipage}{\linewidth}
\renewcommand{\arraystretch}{1.12}
\captionof{table}{MEG datasets and retrieval accuracy with the history-aggregation prior at $\alpha=2.7$.}
\label{tab:datasets}
\smallskip
\centering
\setlength{\tabcolsep}{4pt}
\begin{tabular}{@{}lcccrrcc@{}}
\hline
\multirow{2}{*}{Dataset} & \multirow{2}{*}{Language} & \multirow{2}{*}{Participants} & \multirow{2}{*}{Sensors} & \multirow{2}{*}{Hours} & \multirow{2}{*}{Candidates} & \multicolumn{2}{c}{R@1 (\%)} \\
\cline{7-8}
& & & & & & Local & Fused \\
\hline
MEG-MASC & English & 27 & 208 & 56.2 & 1,464 & 44.42 & 58.82 \\
MOUS & Dutch & 96 & 272 & 80.9 & 825 & 21.72 & 29.98 \\
\hline
\end{tabular}
\end{minipage}
\par\addvspace{\baselineskip}

Contextual priors differ in three respects that our interventions vary separately: the source of history (true preceding words, the decoder's own predictions, or accumulated local scores), the granularity at which prior scores are assigned (sentence or word, shared by all candidates that map to the same entry), and the form of the update (additive fusion or non-additive pruning). Every prior shares one property: candidates that map to the same entry receive the same prior score, so fusion cannot reorder them and the prior acts on competition between entries while local scores act within one.

Our main setting uses a history-aggregation prior that accumulates evidence only from preceding windows. Candidates from the same source sentence form a bucket $\mathcal C_b$, and stimulus timing identifies query sentence presentations and word onsets. Within each presentation, the prior integrates local-score support from earlier windows in word-onset order to infer which bucket matches the query. With $b(j)$ identifying candidate $j$'s bucket, $u_t$ the vector of bucket-size-normalised support from local scores, $H_t$ the accumulated bucket scores, and $\eta$ the update rate,
\begin{equation}
H_t=(1-\eta)H_{t-1}+\eta u_t,
\qquad\pi_{t,j}=\mathcal L(H_{t-1},b(j)),
\label{eq:state}
\end{equation}
where $\mathcal L$ retains up to the four strongest active bucket scores and broadcasts each to that bucket's candidates, giving the prior score vector $\pi_t^{\mathrm{bucket}}$ over distinct buckets. Prediction reads $H_{t-1}$ before the current update; state resets at each presentation, and validation selects $\alpha=2.7$ and $\eta=0.1$.

Language-model (LM) priors replace accumulated support with Qwen3-4B-Base likelihoods~\citep{qwen3technicalreport} at the sentence level (mean per-token log-likelihood per candidate sentence) or the word level (log probability of each candidate's aligned word, shared by candidates carrying that word). Each comes in two history sources: free-running priors condition on preceding predicted words, and teacher-forced priors condition on true preceding words, probing correction when the history is right. For the word-level free-running prior, we generate history once from the frozen decoder's top-1 predictions, yielding one prior-score vector per decoder for the whole sweep. Hard bucket pruning changes the update form: it uses the history-aggregation state to retain four buckets and discard the rest, without adding scores. Appendix~\ref{app:prior} gives the constructions and validation selection.

Sweeps vary only $\alpha$, holding both score streams, history, candidates, and test windows fixed: eight weights from 0.5 to 4.0 for history aggregation, sixteen from 0.05 to 16 for each word-level LM prior. Every weight determines its own repairs and residual errors, against which all three margins are evaluated; cross-intervention comparisons use validation-selected operating points.

\subsection{Confidence estimation and evaluation}

The analyses above use the correct candidate to label outcomes. A system predicting without it must judge two things: whether the output is correct, and whether the prior's revision improved on the local prediction. \emph{Output confidence} targets the first, including whether an output set contains the correct candidate. \emph{Intervention confidence} targets $B(I_t)$, the expected benefit of contextual revision over the local prediction given observable information $I_t$. Under top-1 error loss,
\begin{align}
B(I_t)&=\Pr(\tilde y_t=y_t\mid I_t)-\Pr(\hat y_t=y_t\mid I_t)\notag\\
&=\Pr(\mathrm{repair}\mid I_t)-\Pr(\mathrm{regression}\mid I_t).
\label{eq:benefit}
\end{align}
All probabilities use the same conditioning information. A repair and a prediction that remains correct share the final-correctness label, but only the repair improves on the local prediction. Among initially incorrect windows, final correctness is repair success; across all windows, $B(I_t)$ also depends on preserving correct local predictions.

For output confidence, multilayer perceptrons predict fused top-$K$ success for $K\in\{1,3,5,10,20\}$. Decoder-score estimators combine two local and six fused score features; prior-feature estimators use seven features of prior state, current support, and sentence position; the full estimator combines both. Additional estimators use only the fused margin, the six fused features, or the two local features, with the same hidden-layer architecture and training procedure. Training and calibration use validation data, with cross-validation folds formed by presentation.

These estimates drive an output policy, which selects the smallest $K$ meeting its validation-selected success-probability threshold, or abstains if none qualifies. Emission is the fraction of windows on which a set is output, and coverage is the fraction of emitted sets containing the correct candidate. Validation requires at least 90\% coverage and mean emitted set size at most 4.8. We compare decoder-score and full estimators with a margin-binned reference that applies conformal set calibration~\citep{romano2020classification} in 20 fused-margin bins, reporting emission rate, mean set size, and coverage, also grouped by initial rank (Appendix~\ref{app:output-protocol}).

To compare selection at a fixed output size, fixed-$K$ policies emit their top-$K$ set or abstain, with validation thresholds maximising emission subject to at least 90\% coverage (Appendix~\ref{app:output-size}).

Throughout, estimates are means and standard deviations across the three local decoders, with reach computed within each decoder before averaging; we report a repair-separation cell only when every decoder has at least five repairs, and initial rank serves only as an evaluation label. Uncertainty is quantified at three levels: Wilson intervals describe the quartile repair rates in Figure~\ref{fig:reversal}A, B, which pool evaluated predictions with fixed quartile membership; sentence-cluster bootstrap supports rate contrasts and within-sentence comparisons; and participant bootstrap, sharing each resampled participant set across weights, groups, confidence measures, and decoders, gives the core AUROC intervals and boundary stability (Appendix~\ref{app:stats}).

\section{Results}
\label{sec:experiments}

\subsection{The fused margin reverses its ordering across initial ranks}
\label{sec:exp:reversal}

\begin{figure}[ht]
\centering
\includegraphics[width=\linewidth]{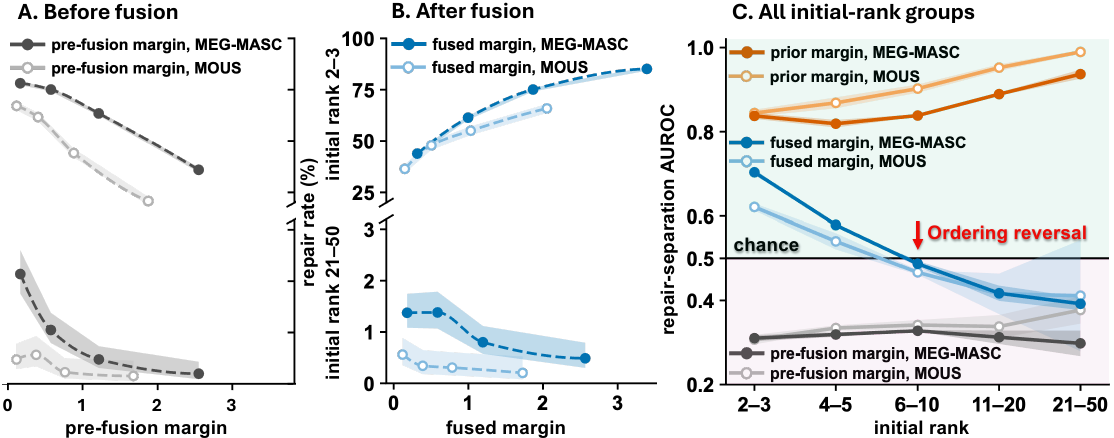}
\caption{\textbf{Confidence ordering reverses as the correct candidate starts lower in the local ranking.} Initially incorrect windows on MEG-MASC (filled) and MOUS (hollow). Panels A and B use the same windows and outcomes, with one point per quartile at its mean margin; the upper and lower rows show initial ranks 2--3 and 21--50, respectively, with different vertical scales. \textbf{(A) Before fusion.} Repair rate decreases with margin in both groups. \textbf{(B) After fusion.} From the smallest to largest quartile, repair rate rises at ranks 2--3 and falls at ranks 21--50. \textbf{(C) All initial-rank groups.} Repair-separation AUROC of the fused, pre-fusion, and prior margins; above 0.5 favours repairs, below 0.5 residual errors. Bands are Wilson 95\% intervals in A, B, and $\pm$1 SD across three local decoders in C; dashed curves interpolate quartile points. MOUS ranks 21--50 average 13 repairs per decoder; Appendix~\ref{app:phenomenon} reports full rates and participant-level intervals.}
\label{fig:reversal}
\end{figure}

\textbf{Confidence ordering reverses as the correct candidate starts lower in the local ranking.} Pooled correctness AUROC is 0.872 on MEG-MASC (Figure~\ref{fig:reversal}), yet among initially incorrect predictions, repair-separation AUROC falls from 0.704 at ranks 2--3 to 0.392 at ranks 21--50. Pooled over all initially incorrect predictions, AUROC is 0.646, so grouping by initial rank exposes opposite orderings that this pooled comparison does not reveal. For the same outcomes, the pre-fusion margin remains negatively ordered across all five groups, with AUROCs of 0.30--0.33. Fusion changes the ordering to positive near the top, while deeper groups remain negative. Repairs concentrate at low initial ranks, leaving a substantial share of errors deeper in the ranking: on MEG-MASC, residual errors starting beyond rank 20 account for 46.6\% of all post-fusion errors.

\textbf{Repair rates across margin quartiles show the same reversal.} From the smallest to largest fused-margin quartile on MEG-MASC, repair rate rises from 42.4\% to 84.4\% at initial ranks 2--3, but falls from 1.37\% to 0.55\% at ranks 21--50. Sentence-cluster bootstrap supports both endpoint contrasts (Appendix~\ref{app:phenomenon}). Rates also rise at ranks 4--5 and fall at ranks 11--20 under this contrast; at ranks 6--10, the interval for the highest-to-lowest quartile ratio includes one.

\textbf{The reversal replicates on MOUS and at the participant level.} There, repair-separation AUROC falls from 0.62 at ranks 2--3 to 0.41 at ranks 21--50, and participant-level intervals support positive ordering near the top and negative ordering further down on both datasets. On MOUS, the 6--10 and 11--20 intervals lie below 0.5, while ranks 21--50 average 13 repairs per decoder and have an interval crossing 0.5 (Appendix~\ref{app:participant-boot}).

\textbf{The reversal persists within sentences, among repairable errors, and for other confidence scores.} Negative fused-margin ordering persists within sentence presentations and among errors whose correct candidate already leads its sentence bucket, the subset an additive prior can in principle repair (Table~\ref{tab:within-sentence}; Appendix~\ref{app:repairable}). Entropy, sentence-score mass gap, and concentration also order repairs negatively (0.36, 0.46, 0.43 at ranks 21--50; Appendix~\ref{app:pooled-output}).

\subsection{A score-level account of margin formation}
\label{sec:exp:hypothesis}

\textbf{A prior can make an error more confident while moving its correct candidate up.} In Table~\ref{tab:examples}, the prior advances the correct candidate from rank 40 to rank 7 while the margin grows from 0.07 to 3.54 around an incorrect winner. We account for this by comparing how repairs and residual errors form their final leads.

\textbf{A repair's margin shrinks as the deficit it must close grows.} Let $G_t^0=\max_{j\ne y_t}s_{t,j}-s_{t,y_t}$ be the correct candidate's initial score deficit and $\Delta_t^\pi=\max_j\pi_{t,j}-\min_j\pi_{t,j}$ the prior score range. On a repair, comparison with the strongest local competitor gives:
\begin{align}
G_t^0&\leq\alpha\Delta_t^\pi,
\label{eq:bound}\\
m(\tilde s_t)&\leq\alpha\Delta_t^\pi-G_t^0
.
\label{eq:repair-margin-main}
\end{align}
Increasing $\alpha$ relaxes both constraints, allowing repairs to close larger deficits with larger margins.

\textbf{Deep residual errors gain margin between two wrong candidates.} With post-fusion score deficit $\tilde G_t=\max_{j\ne y_t}\tilde s_{t,j}-\tilde s_{t,y_t}$, the fused margin equals $-\tilde G_t$ on a repair and is at most $\tilde G_t$ on a residual error. In the latter case, equality holds when the correct candidate is runner-up, in the absence of ties. The correct candidate's runner-up share among residual errors falls from 75.2\% at initial ranks 2--3 to 2.1\% at ranks 21--50 (Table~\ref{tab:referent}). The margin at deep initial ranks therefore mostly compares two incorrect candidates, which can separate even while the correct candidate remains far behind; this is one way fusion leaves an error uncorrected yet more confident.

\textbf{The bound certifies most reversals under history aggregation but few under the word-level LM.} Negative ordering follows when a residual-error margin exceeds a repair's upper bound in more than half of within-group pairs (Appendix~\ref{app:algebra}). This range-based condition certifies 16 of 21 negative cells (weight and rank-group pairs) under history aggregation and 1 of 29 under the word-level teacher-forced prior, whose full score range can greatly exceed the correct candidate's prior advantage over the strongest local competitor. At $\alpha=0.5$ under this prior, measured AUROC falls from 0.674 at ranks 2--3 to 0.419 at ranks 21--50, while the range bound does not certify this negative ordering (Appendix~\ref{app:bound-check}), so the bound alone does not imply the reversal.

These constraints motivate the hypothesis that increasing the prior's contribution extends positive repair ordering towards larger initial score deficits. We examine this response by varying the fusion weight while holding both score streams fixed.

\subsection{Controlled changes in prior contribution move confidence ordering}
\label{sec:exp:reach}

\begin{figure}[ht]
\centering
\includegraphics[width=1\linewidth]{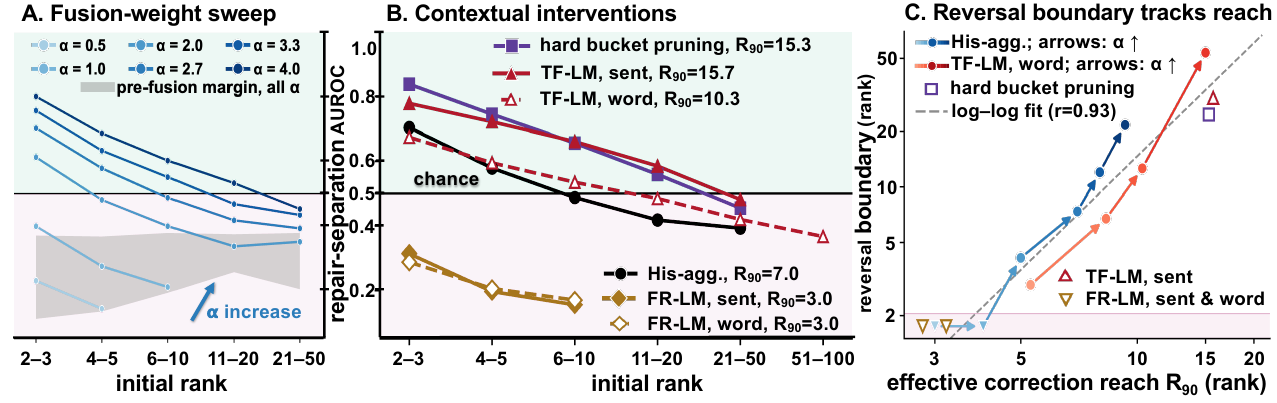}
\caption{\textbf{Changing prior contribution moves the confidence-ordering structure.} MEG-MASC. His-agg.: history aggregation; TF/FR-LM: teacher-forced/free-running LM prior. \textbf{(A) Fusion-weight sweep.} Six weights from the eight-setting history-aggregation sweep, with local and prior scores fixed; grey shows pre-fusion-margin AUROC on the same outcomes. \textbf{(B) Contextual interventions.} Six validation-selected settings: filled markers are sentence-level interventions, hollow markers and dashed curves are word-level LM priors; hard bucket pruning is non-additive. \textbf{(C) Reversal boundary tracks reach.} Arrows follow increasing $\alpha$ for history aggregation and the word-level teacher-forced prior, the latter through $\alpha=0.75$; at higher weights all observed groups through rank 100 are above chance. Open markers correspond to B; the bottom band denotes no positive ordering. Crossings interpolate between group midpoints; the dashed log--log fit summarises the displayed association. Full curves, boundaries, and event counts are in Appendix~\ref{app:mechanism}.}
\label{fig:reach}
\end{figure}

\textbf{We vary only the fusion weight, keeping local and prior scores fixed.} Predictions and outcome labels are recomputed at each weight. Effective correction reach summarises where the resulting repairs begin. At the main history-aggregation setting, 84.8\% of repairs start within the top five ranks and 96.3\% within the top ten, giving $R_{90}=7$ (Figure~\ref{fig:reach}).

\textbf{Raising the history-aggregation weight pushes positive ordering to deeper ranks.} Reach grows from 3.0 to 9.3, and the deepest positively ordered group advances from none through 2--3, 4--5, and 6--10 to 11--20; every increase preserves or advances it, and the sequence is non-decreasing in all 1,000 participant-bootstrap resamples (Appendix~\ref{app:boundary-boot}). On the same outcome labels, the pre-fusion margin stays negatively ordered in all 32 reported cells (0.111--0.378).

\textbf{Positive ordering also extends to larger initial score deficits.} In five equal-mass deficit groups fixed across weights, the deepest positively ordered group advances from none to the fourth group. The 90th percentile of initial deficits among repairs grows from 0.61 to 3.24 score units, while the fifth group remains negatively ordered at $\alpha=4.0$ (Appendix~\ref{app:score-deficit}). Thus positive ordering expands in both initial-rank and score-deficit coordinates under this prior.

\textbf{Under a teacher-forced word-level LM prior, positive ordering continues to expand after accuracy gain peaks.} Between $\alpha=0.05$ and 1, reach grows from 3.0 to 19.3, and positive ordering extends from no group to every observed group through rank 100. Validation selects $\alpha=0.5$ for accuracy; there, positive ordering reaches ranks 6--10, and the gain is 11.20 percentage points. At $\alpha=1$, positive ordering extends further, but the gain falls to 9.16 percentage points.

\textbf{With predicted history, positive ordering appears only after accuracy gain turns negative.} At $\alpha=0.15$, the word-level free-running prior has reach 3.0 and gain 1.25 points, with no positive group; at $\alpha=0.5$, AUROC is 0.539 at ranks 2--3 and below 0.5 elsewhere, while gain is $-0.69$ points. The reversal therefore appears with predicted history as well, although positive ordering emerges only after gain turns negative. At larger weights, sign changes multiply and the deepest positive group retreats even as reach grows (Appendix~\ref{app:word-lm-sweep}).

\textbf{Priors with similar accuracy gains can differ in confidence ordering.} Every prior in Section~\ref{sec:method} shows negative repair ordering, but its extent varies. The sentence-level teacher-forced LM prior and the history-aggregation prior improve accuracy by 14.74 and 14.40 percentage points, respectively, yet retain positive ordering through ranks 11--20 and 4--5. Correction reach captures another aspect of this variation: hard bucket pruning and the sentence-level teacher-forced prior have similar reaches of 15.3 and 15.7, and both retain positive ordering through ranks 11--20. Similar reach can also accompany different ordering boundaries: history aggregation at $\alpha=4.0$ reaches 9.3 with positive ordering through ranks 11--20, whereas the word-level teacher-forced prior at $\alpha=0.5$ reaches 10.3 with positive ordering through ranks 6--10. Prior construction thus shapes both the extent of correction and the ordering of its outcomes.

Nearly half of the remaining errors start beyond rank 20, inside the reversed region; since fusion mixes local and prior support into one margin, we read these two scores separately to identify these errors.

\subsection{Observable information for confidence estimation}
\label{sec:exp:prior}

\textbf{Read before fusion, the prior's own margin separates the outcomes that the fused margin reverses.} The final margin records the winner's lead without distinguishing how local competition and prior support produced it. On the same repairs and residual errors, the prior margin gives 0.937 at ranks 21--50 against 0.392 for the fused margin; under history aggregation it ranges from 0.819 to 0.937 across the five MEG-MASC groups and from 0.84 to 0.99 on MOUS (Figure~\ref{fig:reversal}C), stays above 0.7 in all 32 cells of the fusion-weight sweep while the fused margin reverses (Appendix~\ref{app:boundary-boot}), and among initially correct MEG-MASC predictions it separates those that remain correct from regressions with AUROC $0.777\pm0.022$. Word-level LM priors discriminate less: 0.568--0.759 for repairs and 0.545 for preservation with teacher-forced history ($\alpha=0.5$), and 0.532--0.701 (ranks 2--10) and 0.532 with predicted history ($\alpha=0.15$; Tables~\ref{tab:cross-intervention-auroc} and~\ref{tab:prior-preservation}).

\textbf{Learned estimators rank repairs above residual errors where the fused margin reverses.} Under history aggregation, decoder-score and prior-feature estimators reach repair-separation AUROCs of 0.653 and 0.910 at ranks 21--50 (Table~\ref{tab:confidence-targets}(a)). Among the 10\% of these deep errors scored highest by a prior-state estimator (three prior features), repairs occur at 8.40\%, against 1.80\% for decoder scores and a base rate of 1.05\% (Appendix~\ref{app:pooled-output}). For top-five inclusion, prior-feature discrimination remains strong while decoder-score discrimination is near chance. Across all test windows, the full estimator achieves a correctness AUROC of 0.954, compared with 0.872 for the fused margin, and remains unchanged under any strictly increasing recalibration (Appendices~\ref{app:algebra} and~\ref{app:pooled-output}).

\textbf{Retained scores let selective output emit more, including where errors concentrate.} Emission at matched 92.24\% test-curve coverage (Appendix~\ref{app:feature-output}) rises from 56.7\% with the fused margin to 66.6\% with six fused features, then 72.5\% and 74.5\% as local and prior features are added (Table~\ref{tab:confidence-targets}(b)). For validation-selected policies, adding prior features at ranks 21--50 raises emission from 32.5\% to 38.2\% and coverage from 36.4\% to 50.8\%. At output size one, the full estimator emits at 58.3\% with 90.5\% coverage, versus 48.7\% and 89.1\% for margin binning (Appendix~\ref{app:output-size}).

\begin{table}[t]
\setlength{\abovecaptionskip}{0pt}
\renewcommand{\arraystretch}{1.12}
\centering
\caption{Confidence estimation and selective output with the history-aggregation prior on MEG-MASC; means over three local decoders, with estimator seeds averaged. \textbf{(a)} Top-1 repair and top-5 inclusion AUROC (Table~\ref{tab:learned-strata}). \textbf{(b)} Emission at matched 92.24\% test-curve coverage; $K$ is set size.}
\label{tab:confidence-targets}
\smallskip
\begin{minipage}[t]{0.42\linewidth}
\centering
\textbf{(a) Confidence-estimation AUROC}\par\smallskip
\setlength{\tabcolsep}{3pt}
\begin{tabular}{@{}cccc@{}}
\hline
Initial & Success & Decoder- & Prior- \\
rank & target & score & feature \\
\hline
\multirow{2}{*}{11--20} & Top-1 & 0.716 & 0.848 \\
& Top-5 & 0.552 & 0.752 \\
\cline{2-4}
\multirow{2}{*}{21--50} & Top-1 & 0.653 & 0.910 \\
& Top-5 & 0.522 & 0.867 \\
\hline
\end{tabular}
\end{minipage}\hfill
\begin{minipage}[t]{0.56\linewidth}
\centering
\textbf{(b) Selective output}\par\smallskip
\setlength{\tabcolsep}{3pt}
\begin{tabular}{@{}lccc@{}}
\hline
Features & Dim. & \begin{tabular}[c]{@{}c@{}}Emission\\(\%)\end{tabular} & Mean $K$ \\
\hline
Fused margin & 1 & $56.7\pm0.5$ & 6.3 \\
Fused scores & 6 & $66.6\pm0.7$ & 5.8 \\
Decoder scores & 8 & $72.5\pm0.5$ & 5.2 \\
Full & 15 & $74.5\pm0.4$ & 5.2 \\
\hline
\end{tabular}
\end{minipage}
\vspace{-1em}
\end{table}

\section{Discussion}
\label{sec:discussion}

\textbf{A contextual prior sets both which prediction wins and how confident its errors look.} Larger fused margins make repairs more likely near the top of the local ranking but less likely further down, where nearly half of remaining errors start, while pooled AUROC stays high. Raising the fusion weight moves this reversal to deeper ranks, and under a word-level LM prior it keeps moving after accuracy gain peaks. The prior's own margin, read before fusion, separates repairs from residual errors.

\textbf{The prior can widen the gap between wrong candidates as easily as it lifts the right one.} It adds to every margin only a weighted difference of its own scores, whether or not that difference favours the correct candidate. Confidence ordering therefore follows how the prior reshapes competition, not how much it improves accuracy, and the repair-margin bound holds for any additive fusion. Seeing this requires the local scores, which links confidence analysis to source attribution~\citep{zhang2026audit}: identifying what context changed precedes judging whether the change succeeded.

\textbf{Retained scores restore what recalibration cannot.} Recalibrating the margin preserves its ordering, so no calibration can undo the reversal, and other summaries of the fused scores, such as entropy, reverse as well (Section~\ref{sec:exp:reversal}); keeping local and prior scores separate restores the ordering (Table~\ref{tab:confidence-targets}). A weight chosen for accuracy does not determine confidence, and we can judge a prior by how well its retained scores reveal its correction outcomes, alongside its accuracy gain.

\textbf{Limitations and future work.} Our offline retrieval setting keeps the candidate pool fixed and score streams separate, which makes each correction traceable and the prior's contribution controllable. Free-running histories come from local top-1 rather than fused predictions; teacher-forced histories bracket what richer histories, such as N-best predictions, would provide. Generated candidates would add a changing pool and feedback from fused outputs, and fixed-set assistive interfaces such as BCI spellers~\citep{speier2016integrating} offer a direct test of revision benefit. Our estimators target final correctness; estimating revision benefit, and training objectives that make successful corrections identifiable, are natural next steps. \textbf{Confidence estimation after contextual fusion should therefore retain the local evidence and contextual support behind each prediction, rather than relying on a summary of the fused scores alone.}

\subsection*{Acknowledgements}
The authors acknowledge support from the Vetenskapsrådet under award 2023-00493, the NAISS 2025/22-1173, 2025/23-185, and 2026/3-376.

\clearpage
\subsection*{AI Use Statement}
The authors used ChatGPT and Claude to polish writing and LaTeX formatting. 
The authors produced and verified all analyses, results, and claims and take responsibility for the final content.

\bibliography{iclr2027_conference}
\bibliographystyle{iclr2027_conference}

\clearpage

\appendix

\section{Formal properties of contextual fusion}
\label{app:algebra}

Equation~\ref{eq:poe} multiplies local candidate odds by the corresponding prior odds raised to $\alpha$. For $\alpha>0$, this gives the prior temperature $1/\alpha$, independently of calibration.

\subsection{Bounded corrective capacity}

Let $j^*=\arg\max_{j\ne y_t}s_{t,j}$ be the strongest local competitor on an initially incorrect window. Repair requires
\begin{equation}
G_t^0=s_{t,j^*}-s_{t,y_t}
\leq\alpha(\pi_{t,y_t}-\pi_{t,j^*})
\leq\alpha\Delta_t^\pi,
\end{equation}
where $\Delta_t^\pi=\max_j\pi_{t,j}-\min_j\pi_{t,j}$. Comparison with the same competitor bounds the repaired prediction's final margin:
\begin{align}
m(\tilde s_t)
&\leq\tilde s_{t,y_t}-\tilde s_{t,j^*}\notag\\
&=\alpha(\pi_{t,y_t}-\pi_{t,j^*})-G_t^0\notag\\
&\leq\alpha\Delta_t^\pi-G_t^0.
\label{eq:repair-margin-bound}
\end{align}
For a fixed fusion weight and prior score range, closing a larger deficit leaves less room for the repaired prediction's margin. Ordering additionally depends on the margins of residual errors. Write $z_t=m(\tilde s_t)$ and draw window indices $r$ and $e$ independently and uniformly from repairs and residual errors, respectively, in the same group $\mathcal B$. Then
\begin{equation}
\operatorname{AUROC}_{\mathrm{repair},\mathcal B}(z)
\leq 1-\Pr\!\left[z_e>\alpha\Delta_r^\pi-G_r^0\right].
\label{eq:reversal-sufficient}
\end{equation}
The event on the right makes the residual-error margin exceed the repair margin. If its probability exceeds one half, repair-separation AUROC is below one half under the tie convention in Eq.~\ref{eq:auroc}. The sufficient condition thus connects the individual repair bound to the distribution of residual-error margins.

\subsection{Margin interpretation and recalibration}

On a repair, the correct candidate wins, so
\begin{equation}
m(\tilde s_t)=\tilde s_{t,y_t}-\max_{j\ne y_t}\tilde s_{t,j}=-\tilde G_t.
\end{equation}
On a residual error, let $k$ and $\ell$ index the two largest fused scores. Then
\begin{equation}
m(\tilde s_t)=\tilde s_{t,k}-\tilde s_{t,\ell}
\leq\tilde s_{t,k}-\tilde s_{t,y_t}=\tilde G_t.
\end{equation}
Without ties, equality holds exactly when the correct candidate is runner-up. When the correct candidate is below the top two, the margin compares two incorrect candidates.

For a strictly increasing function $f$, $z_r>z_e$ if and only if $f(z_r)>f(z_e)$. Recalibration by $f$ thus preserves every pairwise ordering and leaves AUROC unchanged. A non-decreasing function can introduce ties but cannot reverse an individual pair. Re-estimating confidence from additional score features can change the ordering, as evaluated in Appendix~\ref{app:readouts}.

\section{Datasets, retrieval task, and local decoder}
\label{app:data}

MEG-MASC supplies 71,736 test windows in 6,027 sentence presentations, scored against 1,464 audio candidates from 123 sentences; MOUS uses an 825-candidate pool. Dataset sizes are in Table~\ref{tab:datasets}. Stimulus timing identifies query presentations, and source metadata identifies candidate sentence buckets. Presentations define prior resets and sentence-resampling units.

We split stimulus identities 70/10/20 into training, validation, and test sets, assigning every occurrence to the same split. MEG-MASC training and validation windows overlapping test windows from the same audio file are removed. We deduplicate candidates by stimulus identity and window index. Each neural window spans $-0.35$ to $+2.65$ seconds relative to word onset.

The neural encoder combines spatial sensor mixing, participant-specific adaptation, and temporal convolutions to predict frozen wav2vec2-large-xlsr-53 audio representations~\citep{baevski2020wav2vec,conneau2021unsupervised}. Scores are Frobenius inner products with normalised audio targets. Training uses matched-window positives and in-batch negatives, with early stopping on validation loss. The three independently trained local decoders provide cached full-pool scores shared across all interventions.

\section{Contextual interventions and operating-point selection}
\label{app:prior}

\subsection{History aggregation and pruning}

At the selected history-aggregation setting, let $\mathcal T_t$ contain the 64 highest-scoring local candidates and $\theta_t=Q_{0.94}(s_t)$. Each sentence bucket receives its largest positive score excess, normalised by the square root of its size:
\begin{equation}
u_t(b)=\frac{\max\bigl(\{(s_{t,j}-\theta_t)_+:j\in\mathcal T_t\cap\mathcal C_b\}\cup\{0\}\bigr)}{\sqrt{|\mathcal C_b|}}.
\label{eq:bucket-support}
\end{equation}
Here $(v)_+=\max(v,0)$. Equation~\ref{eq:state} accumulates this support with $\eta=0.1$. A bucket is active when $H_{t-1}(b)>0$, so $\mathcal L(H_{t-1},b)=H_{t-1}(b)$ for the four highest-scoring active buckets and $0$ otherwise; since state values are non-negative, $\Delta_t^\pi=\max_b H_{t-1}(b)$. Prior margin is the top-two gap of $H_{t-1}$ over all buckets, with inactive buckets entering as zero. Fewer than four buckets are active only at the first window of a presentation, where no candidate is boosted, and hard pruning retains every candidate. Prediction reads the preceding state before the current update and uses $\alpha=2.7$, selected on validation. State resets at each presentation; support comes from preceding windows in word-onset order.

Hard bucket pruning uses the same state but keeps only candidates in its four leading buckets:
\begin{equation}
\tilde s^{\mathrm{mask}}_{t,j}=\begin{cases}
s_{t,j},&b(j)\in\operatorname{TopIdx}_4(H_{t-1}),\\
-\infty,&\text{otherwise},
\end{cases}
\label{eq:hard-mask}
\end{equation}
where $\operatorname{TopIdx}_4$ returns the four highest-scoring bucket indices. The retained local scores and state update are unchanged.

\subsection{Language-model priors}
\label{app:word-lm}

Qwen3-4B-Base scores either candidate sentences by mean per-token log-likelihood or aligned candidate words by
\[
\pi_{t,j}=\log\Pr_{\mathrm{LM}}(w_j\mid h_{t-1}).
\]
Teacher-forced histories contain true preceding words; free-running histories contain decoded preceding words. Histories end before the current word and reset at presentation boundaries. In the word-level free-running sweep, each frozen local decoder supplies one predicted history and one prior-score vector per window for all weights. Changing $\alpha$ therefore changes only the combination of the fixed local and prior scores.

The sentence-level free-running and teacher-forced weights are 0.465 and 1.86. Word-level weights maximise mean validation accuracy gain across decoders, selecting 0.15 and 0.5, respectively. Both word-level priors are swept over sixteen weights from 0.05 to 16. Candidates carrying the same word share its score; we compute the prior margin over distinct words before assigning it to candidates.

\section{Statistical and selection protocols}
\label{app:stats}

Outcome labels are recomputed at every fusion weight, and all confidence measures use those same labels. Thus, fixed pre-fusion and prior scores can have changing AUROCs. Initially incorrect presentation-start windows, where no bucket is active and $\pi_t=0$, are retained as residual errors with prior margin 0; excluding them lowers prior-margin AUROC by at most 0.05 (ranks 2--3) and changes fused-margin AUROC by at most 0.004. We compute estimates within each local decoder and then average them; learned estimators first average their five training seeds within a decoder. A repair-separation cell is omitted if any decoder has fewer than five repairs. Reach uses the interpolated 90th percentile within each decoder. Boundary summaries report the deepest group with mean AUROC above 0.5; full curves retain non-contiguous positive groups.

Sentence-cluster bootstrap resamples presentations, and participant-cluster bootstrap resamples 27 MEG-MASC or 96 MOUS participants, each with 1,000 replicates and 95\% percentile intervals. Each participant sample is shared across weights, rank groups, confidence measures, and decoders; its statistic is the three-decoder mean. Quartile analyses retain their original group membership and pool repeated decoder evaluations of each window. Wilson intervals~\citep{wilson1927probable} describe quartile proportions; sentence bootstrap~\citep{efron1993introduction} quantifies Q4/Q1 contrasts. Within-sentence AUROC pairs repairs with residual errors from the same presentation. Estimator targets and output-selection protocols are given in Appendix~\ref{app:readouts}.

\section{Evidence for confidence-ordering reversal}
\label{app:phenomenon}

\subsection{Quartile repair rates}

Repair rates are higher in the largest than in the smallest fused-margin quartile at initial ranks 2--3 and 4--5, and lower at ranks 11--20 and 21--50 (Table~\ref{tab:fused-quartiles-all}). The Q4/Q1 interval contains one at ranks 6--10, placing the transition between groups 4--5 and 11--20 under this contrast. Pre-fusion-margin repair rates decrease across quartiles in all five groups on the same outcome labels.

\begin{table}[t]
\renewcommand{\arraystretch}{1.12}
\centering
\caption{Repair rates across initial-rank groups when initially incorrect predictions are grouped by (a) fused margin or (b) pre-fusion margin. Q1 and Q4 contain the smallest and largest margins. Q4/Q1 divides their unrounded repair rates, and its intervals are sentence-cluster bootstrap 95\% intervals. Panel (b) uses the same evaluated predictions and fusion-outcome labels as (a).}
\label{tab:fused-quartiles-all}
\par\normalsize
\setlength{\tabcolsep}{4pt}
\textbf{(a) Fused margin}\par\smallskip
\begin{tabular}{lrrrrrrc}
\hline
\multirow{2}{*}{\shortstack{Initial\\rank}} & \multirow{2}{*}{Predictions} & \multicolumn{4}{c}{Repair rate (\%)} & \multicolumn{2}{c}{Q4/Q1} \\
\cline{3-6}\cline{7-8}
& & Q1 & Q2 & Q3 & Q4 & Ratio & 95\% interval \\
\hline
2--3 & 32,277 & 42.44 & 59.75 & 72.61 & 84.39 & 1.988 & [1.933, 2.047] \\
4--5 & 13,725 & 30.80 & 37.60 & 43.57 & 48.18 & 1.564 & [1.467, 1.668] \\
6--10 & 16,861 & 20.61 & 22.85 & 22.16 & 19.15 & 0.929 & [0.842, 1.022] \\
11--20 & 15,244 & 8.27 & 7.43 & 6.32 & 3.73 & 0.451 & [0.357, 0.547] \\
21--50 & 16,613 & 1.37 & 1.47 & 0.77 & 0.55 & 0.404 & [0.223, 0.654] \\
\hline
\end{tabular}
\par\medskip
\textbf{(b) Pre-fusion margin}\par\smallskip
\begin{tabular}{lrrrrrc}
\hline
\multirow{2}{*}{\shortstack{Initial\\rank}} & \multicolumn{4}{c}{Repair rate (\%)} & \multicolumn{2}{c}{Q4/Q1} \\
\cline{2-5}\cline{6-7}
& Q1 & Q2 & Q3 & Q4 & Ratio & 95\% interval \\
\hline
2--3 & 78.34 & 75.47 & 66.32 & 39.05 & 0.498 & [0.481, 0.514] \\
4--5 & 56.44 & 48.70 & 37.02 & 17.98 & 0.319 & [0.293, 0.345] \\
6--10 & 32.12 & 27.71 & 18.27 & 6.67 & 0.208 & [0.181, 0.233] \\
11--20 & 11.34 & 8.53 & 4.38 & 1.50 & 0.132 & [0.097, 0.172] \\
21--50 & 2.14 & 1.16 & 0.60 & 0.27 & 0.124 & [0.061, 0.209] \\
\hline
\end{tabular}
\end{table}

\subsection{Repair distributions and confidence ordering}

The two datasets have similar repair distributions: 67.2\% of MEG-MASC repairs and 65.3\% of MOUS repairs begin at ranks 2--3, while each assigns 0.6\% to ranks 21--50. On MEG-MASC, 84.8\% of repairs begin within the top five and 96.3\% within the top ten, giving $R_{90}=7$. Table~\ref{tab:megmasc-auroc} reports the confidence measures with decoder-level variation and participant-level intervals.

\begin{table}[t]
\renewcommand{\arraystretch}{1.12}
\centering
\caption{Repair-separation AUROC at the main operating point. Point estimates are means across three local decoders. For the MEG-MASC margins, $\pm$ denotes across-decoder SD. Brackets give participant-cluster bootstrap 95\% intervals, sharing each resample across decoders (27 participants on MEG-MASC, 96 on MOUS). Entropy-based confidence and repair counts are means.}
\label{tab:megmasc-auroc}
\par\normalsize
\setlength{\tabcolsep}{4pt}
\textbf{(a) MEG-MASC}\par\smallskip
\begin{tabular}{@{}lcccc@{}}
\hline
\shortstack{Initial\\rank} & \shortstack{Fused\\margin} & \shortstack{Pre-fusion\\margin} & \shortstack{Prior\\margin} & \shortstack{Fused top-16\\entropy-based\\confidence} \\
\hline
\multirow{2}{*}{2--3} & $0.704\pm0.006$ & $0.310\pm0.007$ & $0.838\pm0.007$ & \multirow{2}{*}{0.67} \\
 & [0.687, 0.718] & [0.301, 0.321] & [0.827, 0.847] &  \\
\cline{2-4}
\multirow{2}{*}{4--5} & $0.579\pm0.001$ & $0.319\pm0.006$ & $0.819\pm0.009$ & \multirow{2}{*}{0.55} \\
 & [0.559, 0.596] & [0.308, 0.329] & [0.812, 0.826] &  \\
\cline{2-4}
\multirow{2}{*}{6--10} & $0.487\pm0.007$ & $0.328\pm0.002$ & $0.838\pm0.003$ & \multirow{2}{*}{0.47} \\
 & [0.464, 0.505] & [0.320, 0.335] & [0.828, 0.849] &  \\
\cline{2-4}
\multirow{2}{*}{11--20} & $0.417\pm0.018$ & $0.312\pm0.016$ & $0.889\pm0.004$ & \multirow{2}{*}{0.38} \\
 & [0.397, 0.439] & [0.296, 0.329] & [0.881, 0.898] &  \\
\cline{2-4}
\multirow{2}{*}{21--50} & $0.392\pm0.015$ & $0.298\pm0.031$ & $0.937\pm0.011$ & \multirow{2}{*}{0.36} \\
 & [0.351, 0.435] & [0.271, 0.327] & [0.918, 0.956] &  \\
\hline
\end{tabular}
\par\medskip
\textbf{(b) MOUS}\par\smallskip
\begin{tabular}{@{}lcccc@{}}
\hline
\shortstack{Initial\\rank} & \shortstack{Fused\\margin} & \shortstack{Pre-fusion\\margin} & \shortstack{Prior\\margin} & \shortstack{Repairs per\\local decoder} \\
\hline
\multirow{2}{*}{2--3} & 0.621 & 0.307 & 0.844 & \multirow{2}{*}{1,438} \\
 & [0.600, 0.640] & [0.295, 0.320] & [0.831, 0.856] &  \\
\cline{2-4}
\multirow{2}{*}{4--5} & 0.540 & 0.334 & 0.868 & \multirow{2}{*}{405} \\
 & [0.514, 0.565] & [0.317, 0.350] & [0.853, 0.882] &  \\
\cline{2-4}
\multirow{2}{*}{6--10} & 0.467 & 0.342 & 0.902 & \multirow{2}{*}{267} \\
 & [0.440, 0.495] & [0.322, 0.363] & [0.888, 0.916] &  \\
\cline{2-4}
\multirow{2}{*}{11--20} & 0.416 & 0.338 & 0.952 & \multirow{2}{*}{79} \\
 & [0.383, 0.450] & [0.305, 0.371] & [0.940, 0.963] &  \\
\cline{2-4}
\multirow{2}{*}{21--50} & 0.411 & 0.377 & 0.990 & \multirow{2}{*}{13} \\
 & [0.304, 0.523] & [0.282, 0.457] & [0.985, 0.994] &  \\
\hline
\end{tabular}
\end{table}

\subsection{Within-sentence comparison}

Pairing repairs with residual errors from the same presentation retains the reversal (Table~\ref{tab:within-sentence}). Above initial rank 10, fused-margin AUROC changes from 0.415 across presentations to 0.344 within presentations; prior-margin AUROC remains positive in all five primary groups.

\begin{table}[!ht]
\renewcommand{\arraystretch}{1.12}
\centering
\caption{Within-sentence repair-separation AUROC. Chance is exactly 0.5; 11+ pools all initial ranks above 10.}
\label{tab:within-sentence}
\par\normalsize
\setlength{\tabcolsep}{4pt}
\begin{tabular}{lcccccc}
\hline
Confidence measure & 2--3 & 4--5 & 6--10 & 11--20 & 21--50 & 11+ \\
\hline
Fused margin & 0.672 & 0.520 & 0.404 & 0.357 & 0.284 & 0.344 \\
Pre-fusion margin & 0.284 & 0.276 & 0.256 & 0.242 & 0.204 & 0.268 \\
Prior margin & 0.757 & 0.686 & 0.644 & 0.658 & 0.719 & 0.694 \\
Comparison pairs & 6,457 & 1,279 & 1,308 & 326 & 58 & 1,097 \\
\hline
\end{tabular}
\end{table}

\subsection{Remaining errors and participant uncertainty}
\label{app:participant-boot}

Across the three local decoders, 41,508 evaluated predictions have an initial rank above 20. Fusion repairs 179, leaving 41,329 residual errors: 46.6\% of all post-fusion errors. On this fixed subset, fused-margin repair-separation AUROC is 0.385 with sentence-cluster bootstrap interval $[0.347,0.425]$, compared with 0.950 $[0.934,0.965]$ for prior margin. Pre-fusion margin and fused entropy-based confidence yield 0.305 and 0.365, respectively.

Participant-cluster intervals identify positively and negatively ordered groups on both datasets (Table~\ref{tab:megmasc-auroc}). On MOUS, intervals for ranks 6--10 and 11--20 are below 0.5; ranks 21--50 contain 10--18 repairs per decoder and have a wider interval spanning 0.5. Pooling all initial ranks above 20, fused-margin intervals are $[0.342,0.430]$ for MEG-MASC and $[0.314,0.538]$ for MOUS, while prior-margin intervals are $[0.933,0.967]$ and $[0.992,0.997]$.

\section{Margin formation and empirical bound tightness}
\label{app:outcomes}

\subsection{Representative outcomes}

\begin{figure}[p]
\centering
\includegraphics[width=\linewidth]{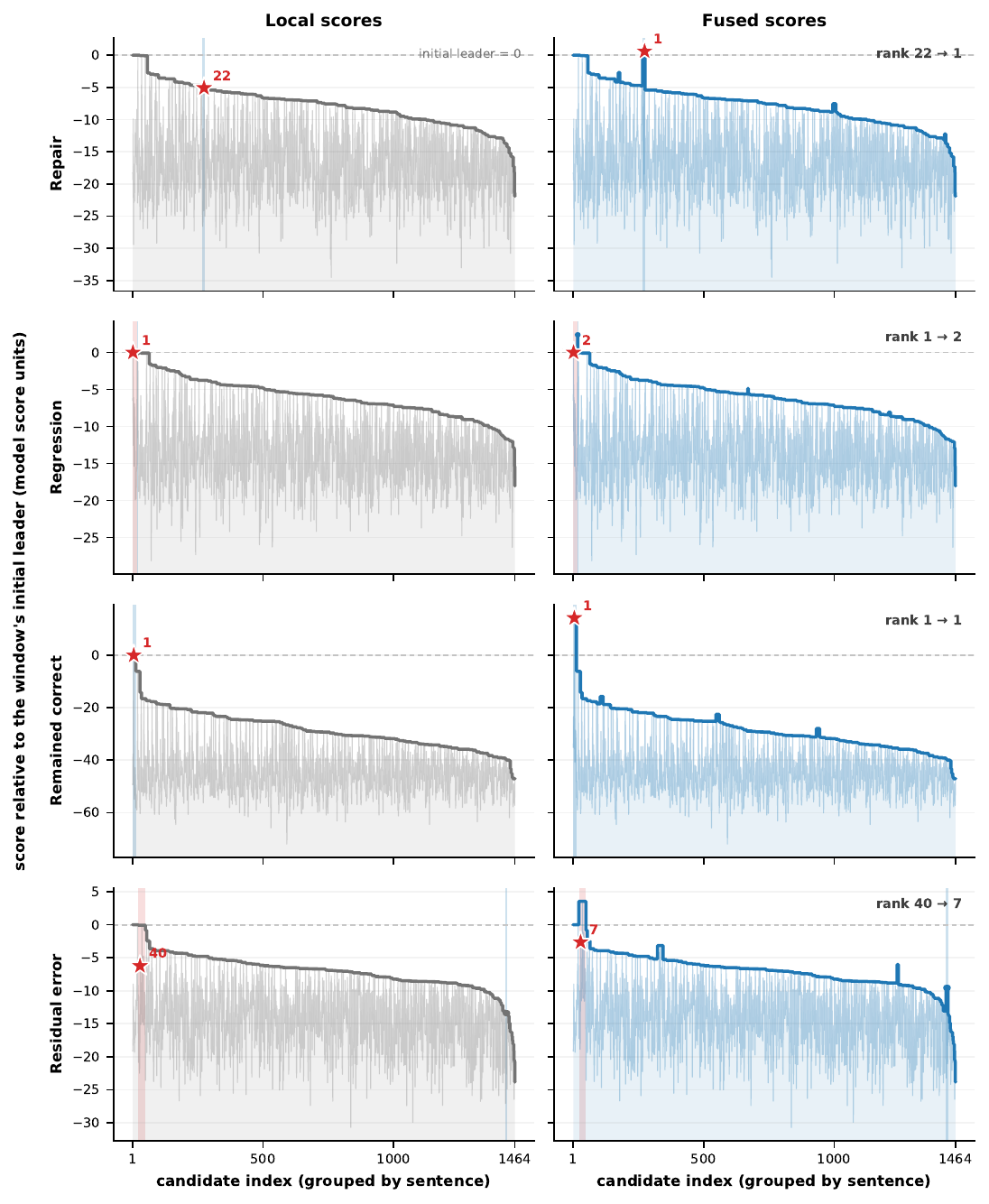}
\caption{\textbf{Candidate scores before and after contextual fusion.} Rows show a repair, a regression, a prediction that remains correct, and a residual error from local decoder seed 0 on MEG-MASC. The left and right columns show local and fused scores. Candidates are grouped by sentence and retain the same horizontal positions within each pair; stars mark the correct candidate. Both score profiles in each row are shifted by the initial local winner's score, preserving ranks and margins. Shaded blocks highlight the contextual and correct-candidate sentence buckets. In the residual example, the correct candidate rises from rank 40 to rank 7 while the prediction remains incorrect.}
\label{fig:outcome-profiles}
\end{figure}

Figure~\ref{fig:outcome-profiles} and Table~\ref{tab:examples} provide score profiles and summary values for the four outcome types.

\begin{table}[htbp]
\renewcommand{\arraystretch}{1.12}
\centering
\caption{Representative MEG-MASC windows from one local decoder.}
\label{tab:examples}
\setlength{\tabcolsep}{4pt}
\begin{tabular}{@{}lcc@{}}
\hline
Outcome & Correct-candidate rank & Margin before $\rightarrow$ after \\
\hline
Repair & $22\rightarrow1$ & $0.05\rightarrow0.61$ \\
Regression & $1\rightarrow2$ & $0.04\rightarrow2.41$ \\
Remained correct & $1\rightarrow1$ & $6.09\rightarrow20.34$ \\
Residual error & $40\rightarrow7$ & $0.07\rightarrow3.54$ \\
\hline
\end{tabular}
\end{table}

\subsection{Correct-candidate participation in the fused margin}

As the initial rank increases, the fused margin increasingly compares two incorrect candidates (Table~\ref{tab:referent}). The difference $\tilde G_t-m(\tilde s_t)$ is the correct candidate's deficit below the runner-up and equals zero when the correct candidate is runner-up.

\begin{table}[htbp]
\renewcommand{\arraystretch}{1.12}
\centering
\caption{Correct-candidate participation on residual errors. Values are mean $\pm$ SD across local decoders. The deficit is measured below the fused runner-up.}
\label{tab:referent}
\setlength{\tabcolsep}{4pt}
\begin{tabular}{@{}lcc@{}}
\hline
Initial rank & Runner-up share (\%) & Median $\tilde G_t-m(\tilde s_t)$ \\
\hline
2--3 & $75.22\pm0.90$ & $0.000\pm0.000$ \\
4--5 & $41.18\pm1.30$ & $0.259\pm0.038$ \\
6--10 & $26.32\pm0.49$ & $0.770\pm0.040$ \\
11--20 & $9.78\pm0.49$ & $1.614\pm0.104$ \\
21--50 & $2.13\pm0.13$ & $2.876\pm0.150$ \\
\hline
\end{tabular}
\end{table}

\subsection{Empirical evaluation of the repair-margin bounds}
\label{app:bound-check}

For repair $r$, let $z_r=m(\tilde s_r)$ and let $j_r^*$ be the strongest local incorrect competitor. Substituting the range bound $\alpha\Delta_r^\pi-G_r^0$ or the competitor bound $c_r=\alpha(\pi_{r,y_r}-\pi_{r,j_r^*})-G_r^0$ into Eq.~\ref{eq:reversal-sufficient} gives AUROC upper bounds $U_{\mathrm{range}}$ and $U_{\mathrm{comp}}$. With measured AUROC $A$, we have $A\leq U_{\mathrm{comp}}\leq U_{\mathrm{range}}$; an upper bound below 0.5 certifies negative ordering.

Table~\ref{tab:bound-summary} summarises all populated cells, each a weight and rank group. Statistics are averaged across decoders, then equally across cells. Under history aggregation, the range bound certifies 16 of 21 negative cells. The other five have AUROCs of 0.434--0.487 and upper bounds of 0.504--0.558, leaving their signs unresolved by this bound.

\begin{table}[htbp]
\renewcommand{\arraystretch}{1.12}
\centering
\caption{Empirical tightness of repair-margin bounds across prior constructions. Gaps average all included cells, which index fusion weights and initial-rank groups. The final column reports cells with mean $U_{\mathrm{range}}<0.5$ among those with mean $A<0.5$.}
\label{tab:bound-summary}
\setlength{\tabcolsep}{4pt}
\begin{tabular}{@{}lrrrr@{}}
\hline
\multirow{2}{*}{Prior} & \multirow{2}{*}{Cells} & \multicolumn{2}{c}{Mean upper-bound gap} & \multirow{2}{*}{\shortstack{Reversals certified\\by range bound}} \\
\cline{3-4}
& & $U_{\mathrm{range}}-A$ & $U_{\mathrm{comp}}-A$ & \\
\hline
History aggregation & 32 & 0.041 & 0.028 & 16/21 \\
Word-level teacher-forced LM & 83 & 0.403 & 0.223 & 1/29 \\
\hline
\end{tabular}
\end{table}

For the word-level teacher-forced prior, the score range spans the 709 distinct words represented in the candidate pool and can greatly exceed the correct candidate's prior advantage over its strongest local competitor. The range bound certifies 1 of the 29 negative cells, while the competitor-based bound certifies 11, all at $\alpha\leq0.5$. The slack of the competitor-based bound is
\[
c_r-z_r=\max_{k\ne y_r}\tilde s_{r,k}-\tilde s_{r,j_r^*}.
\]
The competitor-based bound is attained when $j_r^*$ remains the strongest incorrect competitor after fusion. It is attained more often under history aggregation than under the word-level teacher-forced prior. The empirical tightness therefore depends on both the prior score range and the competitors that determine the final margin.

\section{Correction-reach interventions}
\label{app:mechanism}

\subsection{History-aggregation fusion-weight sweep}
\label{app:boundary-boot}

\begin{figure}[htbp]
\centering
\includegraphics[width=0.85\linewidth]{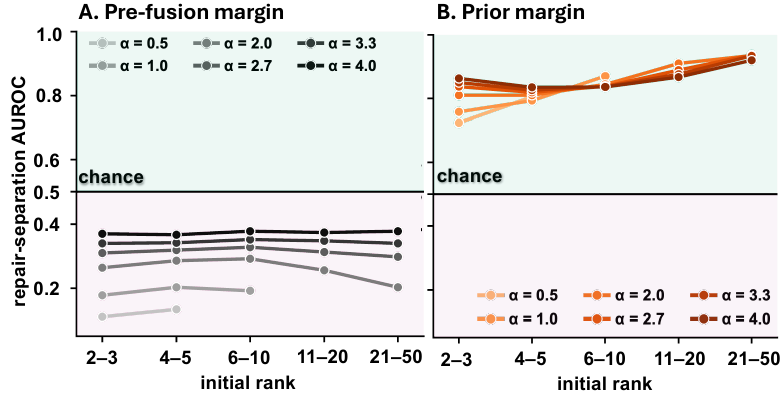}
\caption{\textbf{Pre-fusion and prior margins across the fusion-weight sweep.} Both panels use the history-aggregation prior on MEG-MASC and the same repairs and residual errors at each $\alpha$. Darker curves indicate larger weights; 0.5 marks chance. \textbf{(A) Pre-fusion margin} remains below chance in every reported initial-rank group. \textbf{(B) Prior margin} remains above chance. Groups with fewer than five repairs in any decoder are omitted, as in Table~\ref{tab:alpha-grid}.}
\label{fig:sweep-controls}
\end{figure}

Table~\ref{tab:alpha-grid} reports all eight weights with fixed local scores, prior scores, candidate pool, prior-state trajectories, and test windows. The 32 sufficiently populated cells retain negative pre-fusion ordering (0.111--0.378) and positive prior-margin ordering (0.724--0.937), while the fused margin becomes positive across progressively higher initial-rank groups. Reach increases from 3.0 to 9.3, accompanied by additional repairs and regressions (Table~\ref{tab:alpha-outcomes}). Figure~\ref{fig:sweep-controls} shows the control margins, and Figure~\ref{fig:sweep-deficit} the score-deficit analysis.

The history-aggregation boundary sequence is non-decreasing in all 1,000 participant resamples and matches the point-estimate sequence at every weight in 903 of the 1,000 resamples. 

\begin{table}[htbp]
\renewcommand{\arraystretch}{1.12}
\centering
\caption{History-aggregation sweep: fused-margin repair-separation AUROC (three-decoder means) and deepest positive group. All eight weights are shown. Dashes denote fewer than five repairs in at least one decoder. Control curves are in Figure~\ref{fig:sweep-controls}.}
\label{tab:alpha-grid}
\setlength{\tabcolsep}{4pt}
\begin{tabular}{@{}ccccccc@{}}
\hline
$\alpha$ & 2--3 & 4--5 & 6--10 & 11--20 & 21--50 & \shortstack{Deepest\\positive group} \\
\hline
0.5 & 0.229 & 0.143 & -- & -- & -- & none \\
0.7 & 0.303 & 0.200 & 0.180 & -- & -- & none \\
1 & 0.399 & 0.275 & 0.210 & -- & -- & none \\
1.5 & 0.520 & 0.379 & 0.319 & 0.265 & -- & 2--3 \\
2 & 0.613 & 0.480 & 0.399 & 0.336 & 0.350 & 2--3 \\
2.7 & 0.704 & 0.579 & 0.487 & 0.417 & 0.392 & 4--5 \\
3.3 & 0.758 & 0.633 & 0.551 & 0.468 & 0.434 & 6--10 \\
4 & 0.802 & 0.687 & 0.602 & 0.533 & 0.452 & 11--20 \\
\hline
\end{tabular}
\end{table}
\begin{table}[htbp]
\renewcommand{\arraystretch}{1.12}
\centering
\caption{History-aggregation reach and intervention outcomes at six fusion weights. Counts are per-decoder means.}
\label{tab:alpha-outcomes}
\setlength{\tabcolsep}{4pt}
\begin{tabular}{@{}ccccc@{}}
\hline
$\alpha$ & Repairs & $R_{90}$ & $\Delta\mathrm{R@1}$ (pp) & Regressions \\
\hline
0.5 & 2,376 & 3.0 & 3.30 & 7.3 \\
1.0 & 4,554 & 4.0 & 6.33 & 16.7 \\
2.0 & 8,278 & 5.0 & 11.49 & 34.3 \\
2.7 & 10,381 & 7.0 & 14.40 & 51.3 \\
3.3 & 11,910 & 8.0 & 16.52 & 65.3 \\
4.0 & 13,410 & 9.3 & 18.58 & 88.0 \\
\hline
\end{tabular}
\end{table}

\subsection{Score-deficit groups}
\label{app:score-deficit}

\begin{figure}[htbp]
\centering
\includegraphics[width=0.5\linewidth]{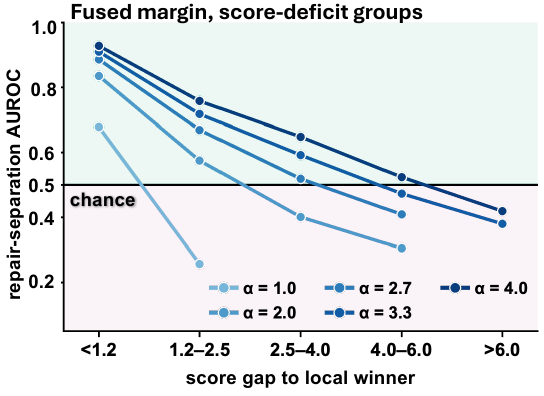}
\caption{\textbf{Fused-margin ordering in score-deficit groups.} History-aggregation prior on MEG-MASC, with the same repairs and residual errors at each $\alpha$ as in Figure~\ref{fig:sweep-controls}. Darker curves indicate larger weights; 0.5 marks chance. Five equal-mass groups of initial score deficit are fixed across weights, with cutoffs 1.24, 2.51, 3.97, and 6.03; the boundary moves toward larger deficits as $\alpha$ increases. The $\alpha=0.5$ curve is omitted because only one group qualifies. Omitted cells follow the event-count rule of Table~\ref{tab:deficit-grid}.}
\label{fig:sweep-deficit}
\end{figure}

Grouping by the initial score deficit $G_t^0=\max_{j\ne y_t}s_{t,j}-s_{t,y_t}$ gives the same direction of boundary movement (Table~\ref{tab:deficit-grid}; Figure~\ref{fig:sweep-deficit}). Five equal-mass groups of initially incorrect predictions remain fixed across weights, with cutoffs 1.24, 2.51, 3.97, and 6.03. Score-deficit reach, $D_{90}=Q_{0.90}(G_t^0\mid\mathrm{repair})$, is computed within each decoder and averaged. It rises from 0.61 to 3.24 while the deepest positive group advances from none to the fourth group.

\begin{table}[t]
\renewcommand{\arraystretch}{1.12}
\centering
\caption{Fused-margin repair-separation AUROC in fixed score-deficit groups on MEG-MASC. Entries are means across three local decoders; dashes denote fewer than five repairs in at least one decoder. Groups 1--5 are ordered by increasing initial score deficit.}
\label{tab:deficit-grid}
\par\normalsize
\setlength{\tabcolsep}{4pt}
\begin{tabular}{ccccccc}
\hline
\multirow{2}{*}{$\alpha$} & \multicolumn{5}{c}{Score-deficit group} & \multirow{2}{*}{\shortstack{Deepest group\\mean AUROC $>0.5$}} \\
\cline{2-6}
& 1 & 2 & 3 & 4 & 5 & \\
\hline
0.5 & 0.441 & -- & -- & -- & -- & none \\
0.7 & 0.561 & 0.148 & -- & -- & -- & 1 \\
1.0 & 0.678 & 0.257 & -- & -- & -- & 1 \\
1.5 & 0.777 & 0.447 & 0.306 & -- & -- & 1 \\
2.0 & 0.835 & 0.575 & 0.401 & 0.305 & -- & 2 \\
2.7 & 0.886 & 0.669 & 0.519 & 0.409 & -- & 3 \\
3.3 & 0.910 & 0.718 & 0.592 & 0.473 & 0.380 & 3 \\
4.0 & 0.928 & 0.758 & 0.648 & 0.524 & 0.419 & 4 \\
\hline
\end{tabular}
\end{table}

\subsection{Structurally repairable errors}
\label{app:repairable}

Shared bucket scores preserve the local order within each bucket: $\tilde s_{t,i}-\tilde s_{t,j}=s_{t,i}-s_{t,j}$ for $i,j\in\mathcal C_b$. An additive repair therefore requires the correct candidate to have the highest local score within its bucket. With $\mathbf{1}[\cdot]$ denoting the indicator function, the structurally repairable subset is
\begin{equation}
\mathrm{repairable}_t=\mathbf{1}\!\left[s_{t,y_t}=\max_{j\in\mathcal C_{b(y_t)}}s_{t,j}\right].
\label{eq:repairable}
\end{equation}
This subset contains 98.6\% of initial errors at ranks 2--3 and 72.5\% at ranks 21--50, retaining every repair. Across the eight-weight sweep, restricting comparisons to this subset changes fused-margin AUROC by at most 0.004 and preserves every boundary in each decoder. At the main operating point, pre-fusion ordering remains negative and prior-margin ordering positive. The reversal therefore persists among errors satisfying the within-bucket condition for repair.

\subsection{Word-level LM sweeps}
\label{app:word-lm-sweep}

Tables~\ref{tab:word-lm-grid} and~\ref{tab:word-lm-controls} report all sixteen weights for each word-level prior with both score channels fixed. The teacher-forced prior extends positive fused-margin ordering from no group at $\alpha=0.05$ to all six observed groups at $\alpha=1$, as reach increases from 3.0 to 19.3. All observed groups remain positive at larger weights, so no crossing is observed through rank 100 for $\alpha\geq1$. Figure~\ref{fig:reach}C plots the observed crossings at $\alpha\leq0.75$.

This extension continues after net accuracy gain peaks: between $\alpha=0.5$ and 1, gain declines from 11.20 to 9.16 percentage points while positive ordering expands from ranks 6--10 to all observed groups. At $\alpha\geq3$, regressions outnumber repairs even though fused-margin repair ordering remains positive throughout the observed range.

\begin{table}[htbp]
\renewcommand{\arraystretch}{1.12}
\centering
\caption{Word-level teacher-forced LM sweep. Fused-margin AUROCs, reach, and accuracy gain are three-decoder means; validation selects $\alpha=0.5$. We show every tested weight and all six rank groups. Dashes follow the repair-count rule. Event counts are in Table~\ref{tab:word-event-counts}.}
\label{tab:word-lm-grid}
\setlength{\tabcolsep}{4pt}
\begin{tabular}{@{}cccccccccc@{}}
\hline
\multirow{2}{*}{$\alpha$} & \multicolumn{6}{c}{Initial rank} & \shortstack{Deepest\\positive group} & \multirow{2}{*}{$R_{90}$} & \shortstack{$\Delta\mathrm{R@1}$\\(pp)} \\\cline{2-7}
 & 2--3 & 4--5 & 6--10 & 11--20 & 21--50 & 51--100 & & & \\
\hline
0.05 & 0.173 & 0.112 & -- & -- & -- & -- & none & 3.0 & +2.21 \\
0.1 & 0.301 & 0.212 & 0.172 & -- & -- & -- & none & 3.0 & +4.20 \\
0.15 & 0.399 & 0.305 & 0.248 & 0.205 & -- & -- & none & 4.0 & +5.88 \\
0.2 & 0.471 & 0.378 & 0.301 & 0.257 & -- & -- & none & 5.0 & +7.29 \\
0.25 & 0.526 & 0.432 & 0.356 & 0.291 & 0.260 & -- & 2--3 & 5.3 & +8.47 \\
0.3 & 0.571 & 0.475 & 0.394 & 0.335 & 0.278 & -- & 2--3 & 6.3 & +9.45 \\
0.4 & 0.633 & 0.547 & 0.480 & 0.417 & 0.356 & 0.267 & 4--5 & 8.3 & +10.65 \\
0.5 & 0.674 & 0.595 & 0.535 & 0.484 & 0.419 & 0.366 & 6--10 & 10.3 & +11.20 \\
0.75 & 0.732 & 0.664 & 0.623 & 0.567 & 0.517 & 0.486 & 21--50 & 15.0 & +10.80 \\
1 & 0.762 & 0.701 & 0.667 & 0.617 & 0.571 & 0.557 & 51--100 & 19.3 & +9.16 \\
1.5 & 0.785 & 0.733 & 0.705 & 0.671 & 0.640 & 0.625 & 51--100 & 25.7 & +5.05 \\
2 & 0.800 & 0.747 & 0.717 & 0.686 & 0.667 & 0.647 & 51--100 & 30.0 & +1.31 \\
3 & 0.806 & 0.765 & 0.733 & 0.703 & 0.687 & 0.682 & 51--100 & 35.7 & -4.19 \\
4 & 0.810 & 0.772 & 0.739 & 0.707 & 0.695 & 0.695 & 51--100 & 39.3 & -7.72 \\
8 & 0.808 & 0.768 & 0.741 & 0.708 & 0.695 & 0.695 & 51--100 & 47.7 & -14.41 \\
16 & 0.796 & 0.750 & 0.727 & 0.690 & 0.680 & 0.677 & 51--100 & 52.7 & -18.55 \\
\hline
\end{tabular}
\end{table}

The free-running prior produces a different trajectory. At its validation-selected weight, $\alpha=0.15$, reach is 3.0 and gain is 1.25 percentage points. Positive ordering first appears at ranks 2--3 when $\alpha=0.5$, where gain is $-0.69$ points. At $\alpha=2$ and 3, ranks 21--50 remain negative while ranks 51--100 are positive; at $ \alpha=4$, the deepest positive group moves back to ranks 11--20. Fusion weight therefore changes both the shape and extent of positive ordering.

\begin{table}[htbp]
\renewcommand{\arraystretch}{1.12}
\centering
\caption{Word-level free-running LM sweep. Fused-margin AUROCs, reach, and accuracy gain are three-decoder means; validation selects $\alpha=0.15$. We show every tested weight and all six rank groups. Dashes follow the repair-count rule. Event counts are in Table~\ref{tab:word-event-counts}.}
\label{tab:word-lm-controls}
\setlength{\tabcolsep}{4pt}
\begin{tabular}{@{}cccccccccc@{}}
\hline
\multirow{2}{*}{$\alpha$} & \multicolumn{6}{c}{Initial rank} & \shortstack{Deepest\\positive group} & \multirow{2}{*}{$R_{90}$} & \shortstack{$\Delta\mathrm{R@1}$\\(pp)} \\\cline{2-7}
 & 2--3 & 4--5 & 6--10 & 11--20 & 21--50 & 51--100 & & & \\
\hline
0.05 & 0.124 & 0.082 & -- & -- & -- & -- & none & 2.0 & +0.64 \\
0.1 & 0.214 & 0.169 & -- & -- & -- & -- & none & 3.0 & +1.03 \\
0.15 & 0.288 & 0.206 & 0.170 & -- & -- & -- & none & 3.0 & +1.25 \\
0.2 & 0.347 & 0.264 & 0.225 & -- & -- & -- & none & 3.0 & +1.28 \\
0.25 & 0.394 & 0.305 & 0.253 & -- & -- & -- & none & 4.0 & +1.20 \\
0.3 & 0.431 & 0.347 & 0.290 & 0.259 & -- & -- & none & 4.0 & +1.00 \\
0.4 & 0.492 & 0.395 & 0.348 & 0.330 & -- & -- & none & 4.0 & +0.27 \\
0.5 & 0.539 & 0.433 & 0.381 & 0.358 & 0.211 & -- & 2--3 & 5.0 & -0.69 \\
0.75 & 0.607 & 0.506 & 0.452 & 0.419 & 0.338 & -- & 4--5 & 5.3 & -3.71 \\
1 & 0.645 & 0.537 & 0.482 & 0.435 & 0.420 & -- & 4--5 & 6.0 & -6.78 \\
1.5 & 0.691 & 0.584 & 0.534 & 0.474 & 0.427 & 0.486 & 6--10 & 7.0 & -12.15 \\
2 & 0.714 & 0.608 & 0.564 & 0.496 & 0.442 & 0.520 & 51--100 & 8.0 & -16.23 \\
3 & 0.739 & 0.644 & 0.592 & 0.527 & 0.451 & 0.519 & 51--100 & 9.0 & -21.49 \\
4 & 0.755 & 0.668 & 0.606 & 0.548 & 0.463 & 0.462 & 11--20 & 9.7 & -24.66 \\
8 & 0.782 & 0.691 & 0.632 & 0.547 & 0.491 & 0.454 & 11--20 & 11.3 & -30.04 \\
16 & 0.787 & 0.690 & 0.626 & 0.529 & 0.487 & 0.463 & 11--20 & 13.0 & -33.13 \\
\hline
\end{tabular}
\end{table}

\begin{table}[htbp]
\renewcommand{\arraystretch}{1.12}
\centering
\caption{Repairs and regressions in the word-level sweeps, shown as means per local decoder.}
\label{tab:word-event-counts}
\setlength{\tabcolsep}{4pt}
\begin{tabular}{@{}crrrr@{}}
\hline
\multirow{2}{*}{$\alpha$} & \multicolumn{2}{c}{Teacher-forced} & \multicolumn{2}{c}{Free-running} \\\cline{2-3}\cline{4-5}
 & Repairs & Regressions & Repairs & Regressions \\
\hline
0.05 & 1,748.0 & 161.0 & 832.3 & 373.3 \\
0.1 & 3,358.3 & 343.3 & 1,536.7 & 796.3 \\
0.15 & 4,771.7 & 553.7 & 2,121.0 & 1,223.3 \\
0.2 & 6,027.0 & 796.7 & 2,597.3 & 1,680.0 \\
0.25 & 7,120.7 & 1,048.0 & 3,000.3 & 2,137.7 \\
0.3 & 8,086.0 & 1,308.7 & 3,323.0 & 2,606.0 \\
0.4 & 9,542.7 & 1,899.3 & 3,783.0 & 3,586.7 \\
0.5 & 10,575.0 & 2,542.3 & 4,058.7 & 4,554.0 \\
0.75 & 11,972.0 & 4,222.0 & 4,303.0 & 6,962.0 \\
1 & 12,434.0 & 5,865.0 & 4,269.0 & 9,131.3 \\
1.5 & 12,378.0 & 8,756.0 & 3,908.7 & 12,627.7 \\
2 & 12,016.7 & 11,078.3 & 3,540.3 & 15,181.3 \\
3 & 11,251.7 & 14,260.7 & 3,035.3 & 18,449.3 \\
4 & 10,659.0 & 16,195.3 & 2,712.3 & 20,399.7 \\
8 & 9,402.0 & 19,742.0 & 2,119.3 & 23,671.7 \\
16 & 8,625.0 & 21,934.7 & 1,783.3 & 25,551.0 \\
\hline
\end{tabular}
\end{table}

\textbf{Pre-fusion and prior margins.}
At each weight, we evaluate the pre-fusion and prior margins on the same repair and residual-error labels as the fused margin, showing how ordering changes for fixed input scores as outcomes change. Through $\alpha=1$, pre-fusion ordering is negative and prior-margin ordering positive in every reported group for both word-level priors (Table~\ref{tab:word-lm-pre}). At larger weights, pre-fusion AUROC approaches 0.5, and some groups become positive. For the teacher-forced prior, ranks 11--20 reach 0.503 at $\alpha=1.5$, with a net accuracy gain of 5.05 percentage points. Teacher-forced prior-margin ordering stays positive throughout the sweep; the free-running prior margin has one negative group, ranks 51--100 at $\alpha=16$ (0.479).

\begin{table}[!ht]
\renewcommand{\arraystretch}{1.12}
\centering
\caption{Pre-fusion and prior-margin repair-separation AUROC in the word-level sweeps. Each range spans the reported three-decoder means over all sufficiently populated initial-rank groups and tested weights in the indicated interval. Both measures use the same cells and outcome labels.}
\label{tab:word-lm-pre}
\par\normalsize
\setlength{\tabcolsep}{4pt}
\begin{tabular}{ccccc}
\hline
\shortstack{Word-level\\LM prior} & Weights & Cells & \shortstack{Pre-fusion\\margin} & \shortstack{Prior\\margin} \\
\hline
\multirow{2}{*}{Teacher-forced} & $\alpha\leq1$ & 47 & 0.102--0.494 & 0.554--0.913 \\
& $\alpha>1$ & 36 & 0.477--0.509 & 0.566--0.706 \\
\cline{2-5}
\multirow{2}{*}{Free-running} & $\alpha\leq1$ & 36 & 0.098--0.475 & 0.522--0.711 \\
& $\alpha>1$ & 36 & 0.450--0.525 & 0.479--0.596 \\
\hline
\end{tabular}
\end{table}

\subsection{Reach and ordering across interventions}
\label{app:reach-ladder}

At the selected operating points, effective correction reach spans 3.0--15.7 across six interventions (Table~\ref{tab:reach-ladder}). Both free-running priors have no positively ordered group; the history-aggregation and word-level teacher-forced priors extend positive ordering through ranks 4--5 and 6--10; hard bucket pruning and the sentence-level teacher-forced prior extend it through ranks 11--20 (Table~\ref{tab:cross-intervention-auroc}).

\begin{table}[t]
\renewcommand{\arraystretch}{1.12}
\centering
\caption{Reach and intervention outcomes at the selected operating points on MEG-MASC. Reach and accuracy gain are mean $\pm$ SD across three local decoders; repair and regression counts are per-decoder means.}
\label{tab:reach-ladder}
\par\normalsize
\setlength{\tabcolsep}{4pt}
\begin{tabular}{lcccc}
\hline
Intervention & $R_{90}$ & Repairs & Regressions & \shortstack{$\Delta\mathrm{R@1}$\\(pp)} \\
\hline
Sentence LM, free-running & $3.0\pm0.0$ & 2,529.0 & 1,579.7 & $1.32\pm0.13$ \\
Word LM, free-running & $3.0\pm0.0$ & 2,121.0 & 1,223.3 & $1.25\pm0.07$ \\
History aggregation & $7.0\pm0.0$ & 10,380.7 & 51.3 & $14.40\pm0.26$ \\
Word LM, teacher-forced & $10.3\pm0.6$ & 10,575.0 & 2,542.3 & $11.20\pm0.23$ \\
Hard bucket pruning & $15.3\pm0.6$ & 16,452.7 & 992.7 & $21.55\pm0.24$ \\
Sentence LM, teacher-forced & $15.7\pm0.6$ & 14,148.3 & 3,573.7 & $14.74\pm0.41$ \\
\hline
\end{tabular}
\end{table}

\begin{table}[htbp]
\renewcommand{\arraystretch}{1.12}
\centering
\caption{Fused- and prior-margin repair-separation AUROC at the six selected settings (three-decoder means). Dashes follow the repair-count rule; blank cells were not assessed. The 51--100 group is used only for word-level priors.}
\label{tab:cross-intervention-auroc}
\setlength{\tabcolsep}{4pt}
\begin{tabular}{@{}llcccccc@{}}
\hline
Prior & Margin & 2--3 & 4--5 & 6--10 & 11--20 & 21--50 & 51--100 \\
\hline
\multirow{2}{*}{Sentence LM, free-running} & Fused & 0.314 & 0.197 & 0.156 & -- & -- &  \\
 & Prior & 0.603 & 0.663 & 0.782 & -- & -- &  \\
\multirow{2}{*}{History aggregation} & Fused & 0.704 & 0.579 & 0.487 & 0.417 & 0.392 &  \\
 & Prior & 0.838 & 0.819 & 0.838 & 0.889 & 0.937 &  \\
\multirow{2}{*}{Hard bucket pruning} & Fused & 0.838 & 0.745 & 0.656 & 0.559 & 0.454 &  \\
 & Prior & 0.773 & 0.713 & 0.675 & 0.638 & 0.621 &  \\
\multirow{2}{*}{Sentence LM, teacher-forced} & Fused & 0.778 & 0.722 & 0.661 & 0.585 & 0.481 &  \\
 & Prior & 0.816 & 0.846 & 0.861 & 0.869 & 0.873 &  \\
\multirow{2}{*}{Word LM, free-running} & Fused & 0.288 & 0.206 & 0.170 & -- & -- & -- \\
 & Prior & 0.532 & 0.560 & 0.701 & -- & -- & -- \\
\multirow{2}{*}{Word LM, teacher-forced} & Fused & 0.674 & 0.595 & 0.535 & 0.484 & 0.419 & 0.366 \\
 & Prior & 0.568 & 0.600 & 0.617 & 0.659 & 0.752 & 0.759 \\
\hline
\end{tabular}
\end{table}

Configurations with similar reach can have different ordering boundaries. History-aggregation fusion at $\alpha=4$ has $R_{90}=9.3$ and positive ordering through ranks 11--20; the word-level teacher-forced prior at $\alpha=0.5$ has $R_{90}=10.3$ and positive ordering through ranks 6--10. Thus, reach locates the repair distribution, while the AUROC curves additionally describe how repair and residual-error margins are arranged.

\section{Prior information and revision outcomes}
\label{app:prior-outcomes}

For the history-aggregation prior, the highest-scoring sentence bucket is correct on 98\% of windows in the highest prior-margin quintile and 20\% in the lowest. Among initially correct predictions, the highest prior-margin region places 99.9\% of correct candidates in the prior's leading bucket. Shuffling candidate-to-sentence assignments changes net additional correct predictions from $10{,}329\pm190$ to $-198\pm23$ per decoder. This contrast links the gain to aligning accumulated support with candidate identity.

Table~\ref{tab:prior-preservation} evaluates prior margin among initially correct test predictions, using those that remain correct as positives and regressions as negatives. At the tested settings, prior margin distinguishes preserved correct predictions from regressions more strongly for history aggregation than for the word-level LM priors.

\begin{table}[htbp]
\renewcommand{\arraystretch}{1.12}
\centering
\caption{Prior-margin AUROC among initially correct MEG-MASC predictions at validation-selected weights. Positives remain correct; negatives are regressions. Means are across three decoders, with SD where available.}
\label{tab:prior-preservation}
\setlength{\tabcolsep}{4pt}
\begin{tabular}{@{}lcc@{}}
\hline
Prior & $\alpha$ & AUROC \\
\hline
History aggregation & 2.7 & $0.777\pm0.022$ \\
Word-level teacher-forced LM & 0.5 & 0.545 \\
Word-level free-running LM & 0.15 & 0.532 \\
\hline
\end{tabular}
\end{table}

\section{Direct confidence measures and learned estimators}
\label{app:readouts}

\subsection{Features and prediction targets}

Table~\ref{tab:learned-features} separates local, fused, and prior information. The prior state summarises preceding windows; current support comes from the present window. Sentence-position features use the window index and the annotated presentation length. Removing position features in separately fitted top-1 estimators changes repair-separation AUROC over local errors from 0.775 to 0.783; removing support entropy as well gives 0.783.

\begin{table}[htbp]
\renewcommand{\arraystretch}{1.12}
\centering
\caption{Observable inputs to the confidence estimators. Decoder scores combine local and fused features (8); prior features combine prior state and additional prior features (7); the full estimator uses all 15.}
\label{tab:learned-features}
\setlength{\tabcolsep}{4pt}
\begin{tabular}{@{}lcp{250pt}@{}}
\hline
Feature group & Count & Inputs \\
\hline
Local & 2 & Pre-fusion margin; local top-16 entropy-based confidence. \\
Fused & 6 & Fused-score gaps at $K\in\{1,3,5,10,20\}$; fused top-16 entropy-based confidence. \\
Prior state & 3 & History indicator; prior-state entropy-based confidence; prior-state margin. \\
Additional prior & 4 & Current-support entropy-based confidence; leading-bucket agreement between prior state and current support; two sentence-position features. \\
\hline
\end{tabular}
\end{table}

Two-layer MLPs with 32 units per hidden layer predict fused top-$K$ inclusion for $K\in\{1,3,5,10,20\}$. Training uses five-fold, presentation-grouped validation data; per-head temperature calibration~\citep{guo2017calibration} uses out-of-fold logits. Repair-focused evaluations use test windows where the local top-$K$ set excludes the correct candidate; at $K=1$, success is repair. Tied candidates share a rank, so rank-based inclusion resolves ties optimistically. Pooled correctness uses all test windows and fused top-1 correctness. Results average five estimator seeds within each of three local decoders.

\subsection{Outcome discrimination}
\label{app:pooled-output}

At initial ranks 21--50, negative repair ordering extends beyond the fused margin: fused top-16 entropy-based confidence, sentence-score mass gap, and sentence-score concentration give AUROCs of 0.36, 0.46, and 0.43. The last two summarise fused score mass between and across sentence buckets. Table~\ref{tab:learned-strata} shows the positive ordering recovered by learned estimates. Over all test windows, the full estimator achieves an AUROC of 0.954, compared with 0.872 for the fused margin; among local errors, the latter achieves 0.646, falling to 0.415 above rank 10 and 0.344 when comparisons are restricted within presentations.

\begin{table}[!ht]
\renewcommand{\arraystretch}{1.12}
\centering
\caption{AUROC for fused top-$K$ success within initial-rank groups on MEG-MASC. Values are means across three local decoders, averaged over estimator seeds within each decoder. The $K=1$ columns measure repair-separation AUROC, and $n_{\mathrm{repair}}$ counts top-1 repairs per local decoder.}
\label{tab:learned-strata}
\par\normalsize
\setlength{\tabcolsep}{4pt}
\textbf{(a) Top-1 success}\par\smallskip
\begin{tabular}{lrrrr}
\hline
\shortstack{Initial\\rank} & $n_{\mathrm{repair}}$ & \shortstack{Local\\scores} & \shortstack{Decoder\\scores} & \shortstack{Prior\\features} \\
\hline
2--3 & 6,971 & 0.399 & 0.891 & 0.720 \\
4--5 & 1,832 & 0.318 & 0.833 & 0.728 \\
6--10 & 1,191 & 0.284 & 0.779 & 0.775 \\
11--20 & 327 & 0.240 & 0.716 & 0.848 \\
21--50 & 58 & 0.271 & 0.653 & 0.910 \\
\hline
\end{tabular}
\par\medskip
\textbf{(b) Top-5 success}\par\smallskip
\begin{tabular}{lrrr}
\hline
\shortstack{Initial\\rank} & \shortstack{Decoder\\scores} & \shortstack{Prior\\features} & Full \\
\hline
6--10 & 0.663 & 0.681 & 0.676 \\
11--20 & 0.552 & 0.752 & 0.676 \\
21--50 & 0.522 & 0.867 & 0.771 \\
\hline
\end{tabular}
\end{table}

Table~\ref{tab:rare-event} complements ranking with rare-repair precision. At ranks 21--50, the prior-state estimator's highest-scoring 10\% of predictions have an 8.40\% repair rate, compared with 1.80\% for decoder scores and a population rate of 1.05\%.

\begin{table}[!ht]
\renewcommand{\arraystretch}{1.12}
\centering
\caption{Rare-repair prediction with learned decoder-score and prior-state estimators on MEG-MASC. P@10\% is the repair rate among the 10\% highest-scoring predictions. The prior-state estimator uses the prior-state group in Table~\ref{tab:learned-features}.}
\label{tab:rare-event}
\par\normalsize
\setlength{\tabcolsep}{4pt}
\begin{tabular}{lrrrrr}
\hline
\multirow{2}{*}{\shortstack{Initial\\rank}} & \multirow{2}{*}{\shortstack{Repair rate\\(\%)}} & \multicolumn{2}{c}{AUPRC} & \multicolumn{2}{c}{P@10\% (\%)} \\
\cline{3-4}\cline{5-6}
& & Decoder-score & Prior-state & Decoder-score & Prior-state \\
\hline
11--20 & 6.44 & 0.129 & 0.396 & 13.70 & 36.40 \\
21--50 & 1.05 & 0.018 & 0.242 & 1.80 & 8.40 \\
\hline
\end{tabular}
\end{table}

\subsection{Selective-output policies}
\label{app:output-protocol}

At the history-aggregation setting $\alpha=2.7$, adaptive policies emit the smallest set whose calibrated success probability exceeds its threshold, or abstain. We select thresholds on validation under coverage $\geq90\%$ and mean $K\leq4.8$. The margin-binned reference applies conformal calibration in 20 fused-margin bins; isotonic calibration~\citep{zadrozny2002transforming} of that margin finds no point satisfying both constraints. Table~\ref{tab:validation-output} reports the selected policies on test windows; validation constraints are not guaranteed on test, where mean $K$ slightly exceeds 4.8, and fixed-$K$ coverage falls below 90\% for $K\geq3$ with learned estimators and at every $K$ for margin-based policies (Table~\ref{tab:fixed-k-output}). Emission uses all windows as its denominator; coverage and mean $K$ use emitted sets.

\begin{table}[htbp]
\renewcommand{\arraystretch}{1.12}
\centering
\caption{Test performance of validation-selected selective-output policies with the history-aggregation prior. Means and SDs are across three local decoders, after averaging estimator seeds within each decoder. The stratified analysis uses these adaptive policies; fixed-$K$ thresholds are selected separately.}
\label{tab:validation-output}
\setlength{\tabcolsep}{4pt}
\begin{tabular}{@{}lccc@{}}
\hline
Policy & Emission (\%) & Coverage (\%) & Mean $K$ \\
\hline
Margin-binned & $55.9\pm8.0$ & $91.7\pm3.0$ & 4.90 \\
Decoder-score & $73.6\pm0.1$ & $91.8\pm0.2$ & $5.02\pm0.05$ \\
Full & $74.5\pm0.8$ & $92.2\pm0.3$ & $5.18\pm0.08$ \\
\hline
\end{tabular}
\end{table}

\label{app:stratified-output}
We apply the same policies to all test windows before grouping by initial rank (Table~\ref{tab:stratified-output}). Adding prior features raises both emission and coverage at ranks 21--50; beyond rank 50, coverage remains low for both policies.

\begin{table}[!ht]
\renewcommand{\arraystretch}{1.12}
\centering
\caption{Selective output by initial rank on MEG-MASC with the history-aggregation prior. Test results at each policy's selected operating point, averaged across local decoders. Emission uses all windows in the group as its denominator; coverage and mean $K$ use emitted sets. Rates are percentages.}
\label{tab:stratified-output}
\setlength{\tabcolsep}{4pt}
\begin{tabular}{@{}lrrrrrr@{}}
\hline
& \multicolumn{3}{c}{Decoder-score} & \multicolumn{3}{c}{Full} \\
\cline{2-4}\cline{5-7}
Initial rank & Emission & Coverage & Mean $K$ & Emission & Coverage & Mean $K$ \\
\hline
1 & 96.8 & 100.0 & 2.36 & 96.9 & 100.0 & 2.33 \\
2--3 & 89.1 & 99.7 & 5.48 & 86.9 & 99.5 & 5.92 \\
4--5 & 72.3 & 97.8 & 9.15 & 70.0 & 97.1 & 8.80 \\
6--10 & 51.7 & 93.1 & 11.20 & 55.6 & 93.5 & 10.32 \\
11--20 & 35.6 & 73.9 & 12.35 & 43.8 & 82.5 & 12.02 \\
21--50 & 32.5 & 36.4 & 12.61 & 38.2 & 50.8 & 12.86 \\
51+ & 30.2 & 3.4 & 12.76 & 30.9 & 7.3 & 13.75 \\
\hline
\end{tabular}
\end{table}

\label{app:output-size}
For fixed-$K$ output, validation instead chooses a separate threshold for each policy and $K$, maximising emission at coverage $\geq90\%$. All searches are feasible; no additional mean-set-size constraint applies. Table~\ref{tab:fixed-k-output} reports test emission and coverage at these thresholds for every assessed $K$.

\begin{table}[htbp]
\renewcommand{\arraystretch}{1.12}
\centering
\caption{Fixed-size selective output on MEG-MASC with the history-aggregation prior. Thresholds are selected on validation separately for each policy and $K$. Entries are test percentages averaged across local decoders, with estimator seeds averaged where applicable. Emission uses all test windows as its denominator; coverage uses emitted sets.}
\label{tab:fixed-k-output}
\setlength{\tabcolsep}{4pt}
\begin{tabular}{@{}llrrrrr@{}}
\hline
Policy & Rate & $K=1$ & $K=3$ & $K=5$ & $K=10$ & $K=20$ \\
\hline
\multirow{2}{*}{Full} & Emission & 58.3 & 71.2 & 77.1 & 86.3 & 99.5 \\
& Coverage & 90.5 & 89.2 & 88.5 & 87.4 & 86.0 \\
\cline{2-7}
\multirow{2}{*}{Decoder-score} & Emission & 56.8 & 69.3 & 74.8 & 84.2 & 99.2 \\
& Coverage & 90.7 & 89.6 & 88.9 & 87.7 & 86.0 \\
\cline{2-7}
\multirow{2}{*}{Margin-binned} & Emission & 48.7 & 55.8 & 63.0 & 76.3 & 94.1 \\
& Coverage & 89.1 & 88.8 & 87.7 & 86.5 & 86.6 \\
\cline{2-7}
\multirow{2}{*}{Margin isotonic} & Emission & 49.7 & 57.9 & 64.7 & 75.9 & 98.4 \\
& Coverage & 88.4 & 87.8 & 87.0 & 86.7 & 86.0 \\
\hline
\end{tabular}
\end{table}

\subsection{Feature comparison at matched operating points}
\label{app:feature-output}

The six feature sets in Table~\ref{tab:feature-output} share cached inputs, a hidden-layer architecture, and the same training procedure. For each run, shifting its five thresholds together traces a test operating curve. Linear interpolation matches emission at 74.52\% or coverage at 92.24\%, the reference full policy's operating values. Runs without a feasible validation point start from the search's penalised optimum. We match per run, then average within and across decoders. These are test-curve comparisons; Table~\ref{tab:validation-output} separately evaluates validation-selected policies.

\begin{table}[htbp]
\renewcommand{\arraystretch}{1.12}
\centering
\caption{Feature comparison with the history-aggregation prior. Matched emission and coverage are percentages; parentheses give mean emitted set size. Validation counts report runs with a feasible point under coverage $\geq90\%$ and mean $K\leq4.8$, before the test-curve sweep. Values are matched per run and then averaged over seeds within decoder; SDs are across decoders.}
\label{tab:feature-output}
\setlength{\tabcolsep}{4pt}
\begin{tabular}{@{}lccc@{}}
\hline
Features & Coverage at & Emission at & Validation \\
& 74.52\% emission & 92.24\% coverage & feasible \\
\hline
Fused margin & $87.2\pm0.1$ (7.1) & $56.7\pm0.5$ (6.3) & 0/15 \\
Fused scores & $89.3\pm0.2$ (5.5) & $66.6\pm0.7$ (5.8) & 6/15 \\
Local scores & $86.4\pm0.1$ (7.7) & $51.9\pm0.9$ (8.2) & 0/15 \\
Prior features & $87.9\pm0.3$ (6.4) & $61.5\pm1.4$ (6.3) & 0/15 \\
Decoder scores & $91.4\pm0.2$ (5.1) & $72.5\pm0.5$ (5.2) & 15/15 \\
Full & $92.2\pm0.2$ (5.2) & $74.5\pm0.4$ (5.2) & 15/15 \\
\hline
\end{tabular}
\end{table}

\end{document}